%% file: main.tex
\documentclass{m3ailpreprint}
\input{premble}

\input{01_title}

\begin{document}
\maketitle

\input{03_introduction}
\input{04_related_work}
\input{05_framework}
\input{06_experiments}
\input{07_conclusions}

\bibliographystyle{plainnat}
\bibliography{references}

\input{08_appendix}

\end{document}

%% file: premble.tex
\usepackage{booktabs}
\usepackage{graphicx}
\usepackage{xspace}

\usepackage{latexsym}
\usepackage[T1]{fontenc}
\usepackage[utf8]{inputenc}
\usepackage{microtype}

\usepackage{amsmath}
\usepackage{amssymb}
\usepackage{bm}

\usepackage{graphicx}
\usepackage{array}
\usepackage{makecell}
\usepackage{booktabs}
\usepackage{multirow}
\usepackage{wrapfig}
\usepackage[table,dvipsnames]{xcolor}

\usepackage{xspace}
\usepackage{soul}

\usepackage{hyperref}
\usepackage{url}
\hypersetup{pdfauthor={}}

\newcolumntype{L}[1]{>{\raggedright\arraybackslash}p{#1}}

\usepackage{enumitem}
\usepackage{multirow}
\usepackage{xcolor}

\renewcommand{\paragraph}[1]{\textbf{#1.}}

%% file: 01_title.tex
\title{Can Agent Harnesses and Inference Engines Hear Each Other? The HEAR Protocol for Agentic LLM Serving}

\author[*]{Jiaqi Zhao}
\author[*]{Haodong Chen}
\author[*,\ddagger]{Jitai Hao}
\author{Wei Zhao}
\author{Jinghao Pang \\}
\author[\dagger]{Qiang Huang}
\author[\dagger]{Jun Yu}

\affiliation{Harbin Institute of Technology (Shenzhen)}

\contribution{$^*$Equal contribution.}
\contribution{$^\ddagger$Project Leader.}
\contribution{$^\dagger$Corresponding authors.}

\abstract{%
  \input{02_abstract}
}

\checkdata[Email]{\email{\{zhaojiaqi, chen.haodong, weizhao, jinghaopang\}@stu.hit.edu.cn}, \\ \email{jitaihao@outlook.com}, \email{huangqiang@hit.edu.cn}, \email{yujun@hit.edu.cn}} 
\checkdata[Repository]{\url{https://anonymous.4open.science/r/hear-3D1E/.}}

%% file: 02_abstract.tex
LLM agents increasingly execute complex workflows involving multi-turn reasoning, tool use, and parallel agents. 
Efficient serving requires decisions that span two layers with complementary information: the agent harness understands workflow dependencies, context lifecycles, and execution objectives, whereas the inference engine observes request queues, KV-cache state, resource pressure, and execution capabilities. Existing interfaces do not systematically connect these views, limiting workflow-aware execution.
We propose \textbf{HEAR}, a bidirectional \textbf{\underline{H}arness--\underline{E}ngine p\underline{A}i\underline{R}ing protocol} for agentic LLM serving. 
HEAR standardizes how the harness communicates workflow intent and execution requirements and how the engine returns runtime state, capabilities, and outcomes. By separating protocol semantics from optimization policies, HEAR supports diverse coordination strategies without changing workflow or model semantics. We instantiate HEAR for online cache-aware runtime coordination and workload-aware execution-mode selection for agent roles.
Across four conversational and research-agent benchmarks under memory-constrained, concurrent serving, HEAR achieves a \textbf{1.61$\times$} batch speedup and reduces median time-to-first-token by \textbf{2.23$\times$} on SCBench. Mooncake shows that workflow intent and live engine state provide complementary benefits across load regimes. On BrowseComp-Plus and DeepResearchBench, workload-specific configurations yield \textbf{1.23$\times$} and \textbf{2.45$\times$} end-to-end speedups, respectively, without observed task-quality degradation. 
These results establish HEAR as a reusable coordination substrate for efficient agentic LLM serving.

%% file: 03_introduction.tex
\section{Introduction}
\label{sec:introduction}


LLM agents tackle complex tasks through multi-round planning, tool use, and collaboration \citep{yao2023react,wu2024autogen}. 
Unlike a single generation, these applications execute workflows of interdependent model calls with evolving contexts.
An agent harness orchestrates the workflow, maintains context, and manages tool interactions, whereas an inference engine, such as vLLM~\citep{kwon2023vllm} or SGLang~\citep{zheng2024sglang}, performs model execution, request batching, and KV-cache management. 
Thus, end-to-end efficiency depends not only on accelerating individual calls, but also on coordinating requests and reusable state across the workflow.

\begin{wrapfigure}{r}{0.48\linewidth}
\centering
\includegraphics[width=0.99\linewidth]{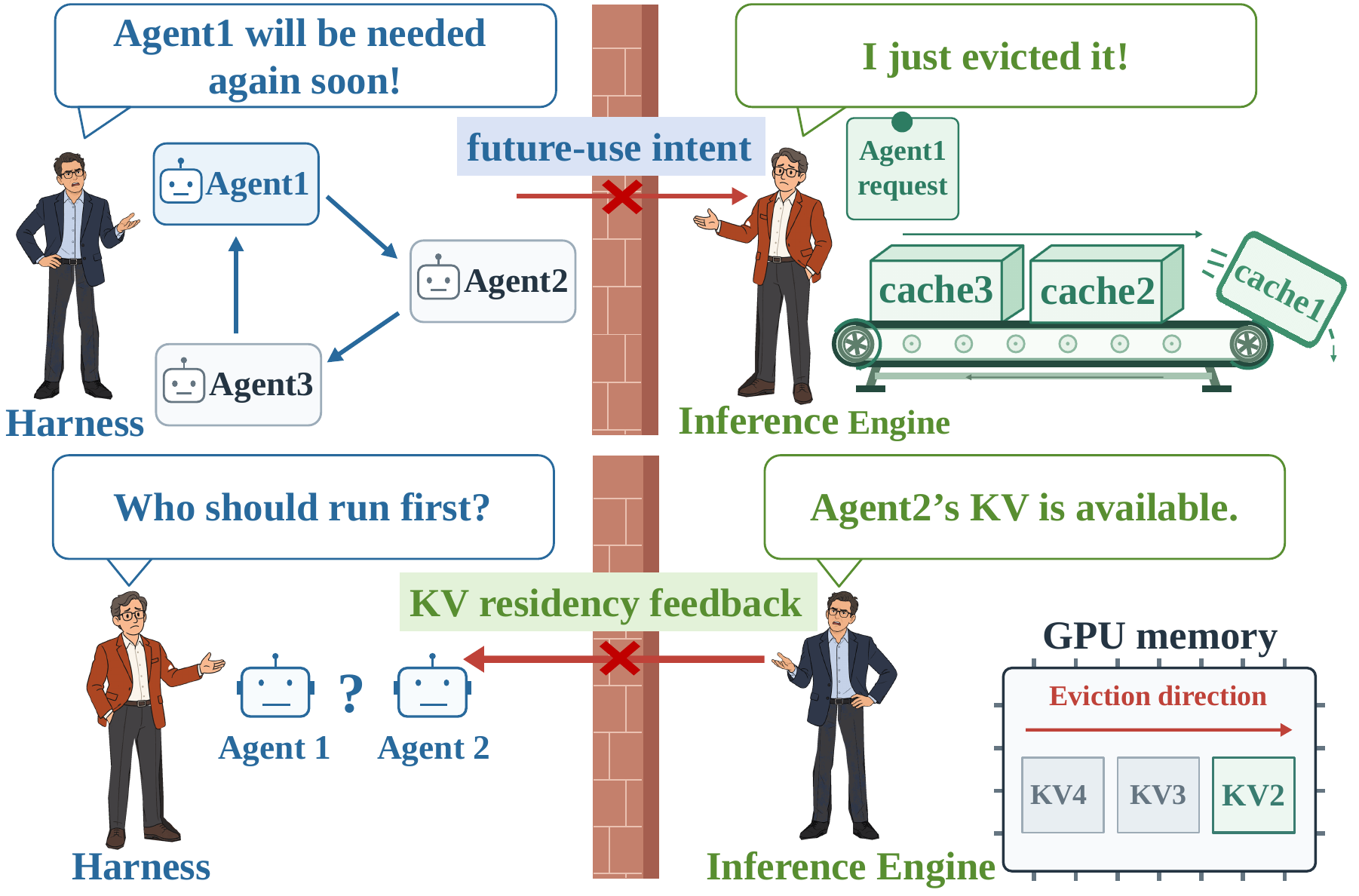}
\vspace{-0.5em}
\caption{\textbf{Motivating example.}
Without workflow intent, the engine may evict KV state shortly before reuse; without KV-residency feedback, the harness may schedule a cold request before an equivalent cache-resident one.}
\label{fig:teaser}
\vspace{-0.5em}
\end{wrapfigure}

This coordination is \emph{difficult} because the information needed for execution decisions is split across the two layers. 
The harness knows request dependencies, context lifecycles, and application objectives, while the engine observes cache residency, request queues, resource pressure, and supported execution modes \citep{pan2025kvflow,chen2026concur}.
Neither view is sufficient in isolation. 
As illustrated in Figure~\ref{fig:teaser}, an engine unaware of future context use may evict KV state immediately before it is needed, forcing costly re-prefilling. 
Conversely, a harness unaware of cache residency may schedule a cold request before a cache-reusable one, even when both are ready and can be legally reordered. 
The same information gap affects execution configuration: selecting an appropriate inference mode requires matching harness-visible workload characteristics and objectives with engine-side efficiency, capacity, and quality trade-offs. 
Effective coordination thus requires a bidirectional interface through which both layers can exchange information and carry out joint decisions.

Prior work demonstrates the value of workflow information and runtime feedback for agentic serving~\citep{lin2024parrot, pan2025kvflow, chen2026concur, yu2026pythia, gim2025pie, zhang2026policyruntime}. 
These efforts reveal a more fundamental systems gap: \textbf{agentic LLM serving lacks a shared control-plane abstraction that connects workflow semantics with runtime execution state.} 
Consequently, individual optimizations must define their own cross-layer signals, control semantics, and operation lifecycles, tightly coupling policies to particular harnesses and inference engines and limiting their portability and composition.

To address this missing abstraction, we introduce \textbf{HEAR}, a bidirectional \textbf{\underline{H}arness-\underline{E}ngine p\underline{A}i\underline{R}ing protocol} for agentic LLM serving. 
HEAR associates each exchange with the relevant request, context version, or serving instance and organizes communication into four semantic categories. 
The harness provides \emph{Execution Description and Intent} and \emph{Execution Requirements and Control}; the engine returns \emph{State and Capabilities} and \emph{Execution Outcomes}. 
HEAR further defines message force and operation lifecycle by distinguishing description from instruction, preference from requirement, observation from guarantee, and acceptance from completion. 
Coordination policies determine what information to exchange and what actions to take, whereas HEAR defines what those exchanges mean and how they should be handled. 
The protocol coordinates how requests from an existing workflow are served without changing the workflow's task-selection logic or the model's semantics.

We demonstrate the generality of HEAR through two complementary policies operating at different timescales. 
\emph{Cache-Aware Runtime Coordination} is an online policy that combines harness-provided waiting constraints and future-reuse intent with engine-reported KV residency and queue state to coordinate request ordering and cache retention. 
\emph{Role-specific inference configuration} is a coarser-grained policy that maps agent roles and their input--output and concurrency characteristics to engine-supported inference modes using measured efficiency, capacity, and quality profiles. 
The two policies use the same interaction contract to answer distinct execution questions: \emph{which ready request should run next} and \emph{how requests from each workflow role should be executed}.

We systematically evaluate HEAR across four conversational and research-agent benchmarks under memory-constrained, concurrent serving. 
HEAR exposes complementary information that neither layer can exploit alone. 
On SCBench~\citep{li2025scbench}, waiting-protected cache-aware scheduling achieves a \textbf{1.61$\times$} batch speedup while reducing median and P95 time-to-first-token by \textbf{2.23$\times$} and \textbf{1.45$\times$}, respectively. 
Mooncake~\citep{qin2026mooncake} shows that future-reuse intent improves cache retention, but must be conditioned on live engine state to remain beneficial under overload. 
On BrowseComp-Plus~\citep{chen2026browsecompplus} and DeepResearchBench~\citep{du2026deepresearch}, workload-specific role configurations deliver \textbf{1.23$\times$} and \textbf{2.45$\times$} end-to-end speedups, respectively, without observed degradation in task quality. 
Together, these results show that harness intent and engine feedback are complementary, workload-dependent signals for execution coordination.

Our contributions are summarized as follows:
\begin{itemize}[nolistsep,left=1pt]
    \item We introduce \textbf{HEAR}, a bidirectional Harness-Engine protocol that associates cross-layer exchanges with requests and contexts, organizes them into four semantic categories, and defines their message force and operation lifecycle.

    \item We instantiate two \textbf{execution-level coordination policies}, online cache-aware runtime coordination and role-specific inference configuration, showing how a common interaction contract supports both runtime scheduling and cache decisions and coarser-grained execution-mode selection.

    \item Through \textbf{controlled evaluations} on SCBench, Mooncake, BrowseComp-Plus, and DeepResearchBench, we show that combining harness intent with engine state improves conversational-serving efficiency and yields \textbf{1.23$\times$--2.45$\times$} end-to-end speedups for research-agent workflows without observed degradation in task quality.
\end{itemize}

%% file: 04_related_work.tex
\section{Related Work}
\label{sec:related_work}

\paragraph{Agent Harnesses and Workflow Orchestration}
Agent harnesses such as AutoGen~\citep{wu2024autogen}, AgentScope~\citep{gao2025agentscope}, LangGraph~\citep{langchain_langgraph}, and Deep Agents~\citep{langchain_deepagents} support multi-agent communication, stateful control flow, model invocation, context management, and tool use.  
LangGraph represents dynamic workflows as stateful graphs, while Deep Agents adds planning and subagent delegation. 
These frameworks determine \emph{what} agents execute and \emph{when} model calls become ready, but lack direct access to runtime conditions independently managed by inference engines, including request queues, batching, KV-cache residency, and resource pressure. 
\textbf{HEAR} bridges this boundary by conveying workflow intent and requirements to the engine and returning engine state and execution outcomes to the harness.

\paragraph{LLM Inference Systems}
LLM inference systems improve request-level efficiency through continuous batching~\citep{yu2022orca}, chunked prefill~\citep{agrawal2024sarathi}, and prefill-decode disaggregation~\citep{zhong2024distserve}.
KV-cache optimizations include PagedAttention for flexible allocation and sharing in vLLM~\citep{kwon2023vllm}, RadixAttention for prefix reuse in SGLang~\citep{zheng2024sglang}, asynchronous multi-tier loading and prefetching in LMCache~\citep{liu2025lmcache,vllm_kv_offloading}, and selective retention, dynamic context selection, or compression in H$_2$O, SnapKV, OmniKV, and KIVI~\citep{zhang2023h2o, li2024snapkv, hao2025omnikv, liu2024kivi}.

These systems provide engine-side mechanisms but do not define how workflow semantics reach the engine or how runtime state and outcomes return to the harness. 
Rather than replacing their schedulers, cache managers, or execution modes, \textbf{HEAR} provides a bidirectional contract for communicating context lifecycles, future reuse, and execution requirements; discovering engine capabilities; requesting supported operations; and observing their outcomes.

\paragraph{Cross-Layer Coordination for Agentic Serving}
Recent systems combine application structure with runtime feedback for targeted optimizations. 
Parrot exposes inter-request dependencies, while Agentix prioritizes calls from program history~\citep{lin2024parrot,luo2026agentix}. 
KVFlow uses workflow structure for retention and prefetching, whereas PBKV predicts dynamic agent invocations~\citep{pan2025kvflow,zheng2026pbkv}. 
InferCept and Continuum preserve KV state across external-operation pauses~\citep{abhyankar2024infercept,li2025continuum}, while CONCUR controls concurrency using cache pressure~\citep{chen2026concur}. 
Broader systems include Helium's query-plan-based caching and scheduling~\citep{wadlom2026helium}, Pythia's predictability-driven caching, scheduling, and autoscaling~\citep{yu2026pythia}, Pie's application-defined inferlets for generation and KV management~\citep{gim2025pie}, and an intermediate policy runtime providing shared identities and observation, scoring, prediction, and action primitives~\citep{zhang2026policyruntime}. 
These designs couple cross-layer signals to particular policies or execution substrates.

\textbf{HEAR} instead standardizes the bidirectional \emph{interaction contract} between an existing harness and engine without prescribing policies, relocating application logic, or adding a runtime tier. 
It associates workflow descriptions and controls with requests and contexts, returns engine state, capabilities, and outcomes, and defines message force and operation lifecycle. 
Policies thus retain their decision rules while sharing reusable semantics for cross-layer coordination.

%% file: 05_framework.tex
\section{The HEAR Protocol}
\label{sec:framework}

\begin{figure}[t]
  \centering
  \includegraphics[width=0.99\textwidth]{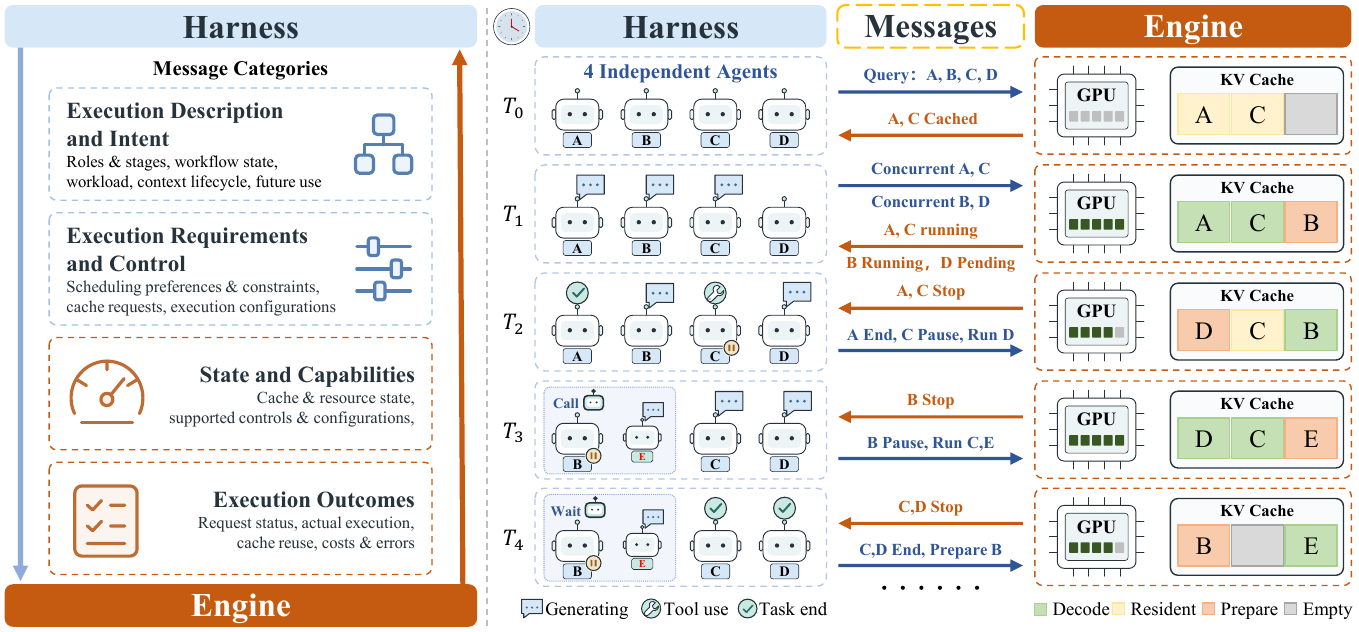}
  \vspace{-0.5em}
  \caption{\textbf{Overview of HEAR and an illustrative dynamic-agent trace.}
  \emph{Left:} HEAR organizes bidirectional harness-engine communication into four semantic message categories.
  \emph{Right:} As the workflow evolves, the harness sends workflow descriptions and execution controls (blue), while the engine returns runtime state and execution outcomes (orange), forming a feedback loop for scheduling and KV-cache management.
  KV-cache colors denote decoding, resident, preparing, and empty states. The trace illustrates one policy instantiation; HEAR does not prescribe this sequence of interactions.}
  \label{fig:framework}
\end{figure}

HEAR is a bidirectional execution-coordination protocol between an agent harness and an inference engine. It defines what cross-layer messages mean and how they are handled, while leaving individual policies to decide what information to exchange and what actions to take. 
Figure~\ref{fig:framework} summarizes the protocol and illustrates how repeated exchanges can coordinate a dynamic workflow as both workflow progress and engine conditions evolve.

\subsection{Scope and Execution Model}
\label{sec:framework:overview}

The harness and engine retain distinct responsibilities: the harness manages workflow semantics, i.e., agent roles, dependencies, readiness, and context lifecycles, while the engine manages model execution, request queues, KV caches, and serving resources.
HEAR connects these views to coordinate request ordering and admission, cache preparation and retention, and supported execution configurations. 
It governs \emph{how} workflow-selected requests are served without changing task selection, relaxing dependencies, or modifying model semantics. 
Workflows may evolve dynamically without being known in advance, and each exchange identifies its associated request, context version, or serving instance, linking workflow requirements to the corresponding engine state and outcomes.

\textbf{Illustrative Trace.}
Figure~\ref{fig:framework} (right) tracks four agents across decision points $T_0$--$T_4$. 
At $T_0$, the harness queries the current requests of agents A--D, and the engine reports resident KV state for A and C. 
As agents execute, pause, finish, or create new work, the harness enforces request validity and dependencies, while engine feedback informs admission, ordering, resumption, and KV preparation.
This trace illustrates one HEAR-enabled policy rather than a protocol-mandated sequence.

\input{tables/main/message_category}

\subsection{Message Categories}
\label{sec:framework:messages}

Cross-layer coordination requires both the conditions motivating a decision and the information or controls needed to carry it out. 
HEAR organizes these exchanges into four semantic categories, summarized in Figure~\ref{fig:framework} (left) and Table~\ref{tab:protocol_messages}.

\paragraph{Harness-to-Engine Messages}
\emph{Execution Description and Intent} conveys execution-relevant workflow facts or predictions, including agent roles, readiness, workload characteristics, context lifecycles, and expected reuse.
\emph{Execution Requirements and Control} requests engine actions, such as scheduling preferences, waiting-time constraints, cache operations, or execution configurations.
The former explains the conditions surrounding a request; the latter asks the engine to act.

\paragraph{Engine-to-Harness Messages}
\emph{State and Capabilities} reports the conditions under which decisions must be made, including cache residency, queue state, resource pressure, supported operations, and available configurations. 
\emph{Execution Outcomes} reports how an inference or control request was handled, including its status, realized configuration, cache reuse, resource cost, or failure. State and Capabilities support decision making, whereas outcomes describe realized effects.

The categories specify semantic roles rather than packet formats. 
A single exchange may contain information from multiple categories, and an implementation may expose only the categories and fields it supports. 
Capability reports allow the harness to discover this supported subset instead of assuming that every engine implements every control.

\subsection{Interaction Semantics}
\label{sec:framework:interactions}

Message categories identify the purpose of an exchange but do not determine its force or lifecycle. HEAR therefore defines four additional distinctions, summarized in Table~\ref{tab:protocol_interactions}.

\input{tables/main/representive_message}

\textbf{Intent vs. Control.}
Descriptions and predictions provide decision context; only explicit control requests ask the receiver to act. For example, reporting that a context is likely to be reused does not itself request that the engine retain or prepare its KV state.

\textbf{Preference vs. Requirement.}
Preferences are best-effort signals, whereas requirements constrain valid execution. Thus, a cache-aware priority may reorder ready requests but cannot violate workflow dependencies or waiting-time protection; an unsatisfied requirement must be rejected or handled through an explicitly permitted fallback.

\textbf{Observation vs. Guarantee.}
State and capability reports describe current conditions and supported operations without reserving resources or guaranteeing future availability. 
For example, A and C's $T_0$ residency may guide scheduling but does not reserve their prefixes.

\textbf{Acceptance vs. Completion.}
Acceptance indicates that an operation has been admitted for processing, whereas completion confirms that its intended effect has been realized. 
In Figure~\ref{fig:framework} (right), requesting preparation of B does not imply that B's KV state is immediately available. The corresponding outcome must distinguish acceptance, completion, rejection, lack of support, and execution failure.

Together, these distinctions prevent policies from treating descriptions as commands, observations as reservations, or accepted operations as completed ones. Later decisions can therefore depend on realized engine effects rather than assumptions about previously issued controls.

\subsection{Policy Integration}
\label{sec:framework:policy}

\paragraph{Joint-Information Policy}
HEAR separates the protocol from the policies that use it. 
Orchestration methods such as Conductor and Puppeteer derive decisions from task and agent state~\citep{nielsen2026conductor, dang2025orchestration}, but application-side information alone does not reveal runtime conditions such as request-level cache residency or queue pressure. 
HEAR makes these engine-side signals available alongside workflow information, enabling policies to condition execution decisions on both views.
We express this joint-information decision as:
\begin{equation}
\label{eq:coordination_policy}
    d_i = \pi(h_i, s_j),    
\end{equation}
where $h_i$ denotes the workflow and request information available at decision point $T_i$, $s_j$ is the latest relevant engine update, and $\pi$ selects an execution arrangement or configuration $d_i$. 
As the two layers evolve independently, the latest applicable engine update $s_j$ need not have been produced at $T_i$.

\paragraph{Illustrative Mapping}
In the illustrative trace, $h_i$ captures which agents are ready, running, paused, or finished, while $s_j$ captures the latest cache-residency and execution feedback. At $T_0$, for example, the workflow view identifies agents A--D, whereas the engine view reports resident KV state for A and C. A policy combines these views to choose an execution decision; subsequent outcomes update the information available to later decisions.

A policy may execute at either layer. It can apply a decision locally or communicate it through the controls defined in Section~\ref{sec:framework:interactions}.
Without engine feedback, Equation~\ref{eq:coordination_policy} reduces to the harness-local decision $d_i=\pi(h_i)$; without workflow information, the engine remains limited to request-local state.

\textbf{Compatibility with Existing Policies.}
HEAR supports policies that already combine application and engine information. For example, PBKV integrates predicted agent invocations with memory and bandwidth constraints for KV prefetching~\citep{zheng2026pbkv}, while Pythia combines workflow progress with queue and capacity state for scheduling~\citep{yu2026pythia}. Their predictors and optimization rules remain unchanged; adapters expose the required workflow descriptions, engine observations, and controls through HEAR. 
Thus, HEAR provides a common semantic substrate rather than a fixed algorithm. Section~\ref{sec:expt} demonstrates this separation at two timescales: online cache-aware coordination on SCBench and Mooncake Bench, and role-specific execution configuration on BrowseComp-Plus and DeepResearchBench.

%% file: tables/main/message_category.tex
\begin{table}[t]
\centering
\small
\renewcommand{\arraystretch}{1.4}
\caption{\textbf{Bidirectional message categories and representative information.}}
\label{tab:protocol_messages}
\vspace{-0.5em}
\begin{tabular}{p{0.25\linewidth} p{0.7\linewidth}}
    \toprule
    \textbf{Category} & \textbf{Representative Contents} \\
    \midrule
    \rowcolor{black!6} 
    \multicolumn{2}{c}{\emph{Harness $\rightarrow$ Engine}} \\
    \textbf{Execution Description and Intent} &
    Agent roles and stages; workflow dependencies and readiness;
    request sizes and output limits; context lifecycles;
    known future uses and explicitly marked predictions. \\
    \textbf{Execution Requirements and Control} &
    Scheduling preferences and waiting-time constraints;
    cache preparation, retention, and release requests;
    allowed or requested execution configurations and serving targets. \\

    \rowcolor{black!6} 
    \multicolumn{2}{c}{\emph{Engine $\rightarrow$ Harness}} \\
    \textbf{State and Capabilities} &
    Reusable prefix ranges and residency tiers;
    allocatable capacity, queue state, and resource pressure;
    supported controls and configurations;
    calibrated or estimated execution costs. \\
    \textbf{Execution Outcomes} &
    Status updates for inference and control requests
    (accepted, completed, rejected, unsupported, or failed);
    actual configuration, permitted fallback, and cache reuse;
    latency, resource use, retries, and errors. \\
    \bottomrule
\end{tabular}
\end{table}

%% file: tables/main/representive_message.tex
\begin{table}[t]
\centering
\small
\setlength{\tabcolsep}{5pt}
\renewcommand{\arraystretch}{1.3}

\caption{\textbf{Key semantic distinctions in bidirectional protocol interactions.}}
\label{tab:protocol_interactions}
\vspace{-0.5em}
\begin{tabular}{p{0.25\linewidth} p{0.7\linewidth}}
\toprule
\textbf{Distinction}
& \textbf{Protocol Interpretation} \\
\midrule

\textbf{Intent vs.\ Control}
& Descriptions expose current or anticipated execution conditions; only explicit control requests ask the receiver to act. \emph{For example, reporting that context $c$ will be} \emph{reused is distinct from requesting that its KV cache} \emph{be retained or prepared.} \\

\addlinespace[0.4em]

\textbf{Preference vs.\ Requirement}
& Preferences are best-effort, whereas requirements constrain valid execution. An unsatisfied requirement must be rejected or handled by an explicitly permitted fallback. \\
& \emph{For example, a priority hint may affect request} \emph{ordering, but it cannot override a mandatory} \emph{protection condition.} \\

\addlinespace[0.4em]

\textbf{Observation vs.\ Guarantee}
& State reports describe observed conditions, and capability reports describe supported operations. Neither reserves resources nor guarantees that a future request will be admitted. \\
& \emph{For example, GPU-residency feedback may guide} \emph{scheduling without reserving the prefix.} \\

\addlinespace[0.4em]

\textbf{Acceptance vs.\ Completion}
& Acceptance means that an operation has beenadmitted for processing; completion confirms that its stated effect has been realized. Rejected, unsupported, and failed operations remain distinct outcomes. \\
& \emph{For example, an accepted cache-preparation} \emph{request may still be pending.} \\

\bottomrule
\end{tabular}
\end{table}

%% file: 06_experiments.tex
\section{Experiments}
\label{sec:expt}

\subsection{Experimental Setup}
\label{sec:experimental_setup}

We evaluate \textbf{HEAR} on two execution decisions that combine harness workflow information with engine state. 
\textbf{Cache-aware runtime coordination} uses rapidly changing cache and queue state to control request ordering and KV retention, evaluated on a controlled closed-loop SCBench workload~\citep{li2025scbench} and a production-derived Mooncake trace~\citep{qin2026mooncake}. 
\textbf{Capability-aware execution configuration} combines workload descriptions with engine capability profiles to select role-specific inference modes, evaluated on BrowseComp-Plus~\citep{chen2026browsecompplus} and DeepResearchBench~\citep{du2026deepresearch} using benchmark-specific workflows built on a common LangGraph harness. All policies preserve the original task-selection logic and workflow dependencies.

Within each setting, all methods use identical models, inputs, prompts, workflow and tool budgets, and hardware. Policies and method-specific parameters are frozen before evaluation, each arm starts from an independently initialized serving state, and reported results include all failures and retries. 
We assess end-to-end efficiency and task quality, using request-, token-, and cache-level measurements for attribution. Implementation details and complete results are provided in
Appendices~\ref{app:implementation}
and~\ref{app:additional_results}.

\subsection{Cache-Aware Runtime Coordination}
\label{sec:cache-sched}

\paragraph{HEAR Instantiation and Runtime Policies}
We instantiate Equation~\ref{eq:coordination_policy} as an online coordination policy. At decision point $T_i$, the harness describes the ready requests $\mathcal{R}_i$, their contexts $\mathcal{C}_i$, and applicable execution requirements $\mathcal{U}_i$, including waiting protection. 
The engine returns the latest relevant cache availability $K_j$ and queue state $Q_j$:
\begin{equation}
\label{eq:scbench_information}
  h_i=(\mathcal{R}_i,\mathcal{C}_i,\mathcal{U}_i), \qquad
  s_j=(K_j,Q_j).
\end{equation}
The index $j$ allows engine feedback to update asynchronously from harness decision point $i$. The scheduling policy produces:
\begin{equation}
\label{eq:scbench_policy}
  d_i = \pi_{\mathrm{sched}}(h_i,s_j) =(\bm{p}_i,\mathcal{U}_i),
\end{equation}
where $\bm{p}_i$ assigns priorities to ready requests. 
Under HEAR, cache and queue reports provide \emph{State and Capabilities}; priorities are scheduling preferences under \emph{Execution Requirements and Control}; and waiting protection is a mandatory requirement. 
The engine applies each preference subject to these requirements and returns the realized execution outcome.

We compare three execution-order policies. \textbf{FCFS} uses the default engine scheduler~\citep{kwon2023vllm}. Inspired by KVFlow~\citep{pan2025kvflow}, \textbf{Cache-Aware} converts engine-reported KV residency into harness-side scheduling preferences; \textbf{Cache-Aware+Guard-$W$} additionally prevents bypass or preemption after a request waits $W$ seconds. We evaluate $W \in \{40, 60\}$ on SCBench and $W=40$ on Mooncake.
For Mooncake, we also evaluate \textbf{Session-Aware}, a KV-retention policy inspired by prediction-guided methods~\citep{zheng2026pbkv, yu2026pythia}. 
After each request, the harness estimates session continuation from its depth and prior inter-turn intervals, explicitly marking the estimate as descriptive intent rather than a retention requirement. The engine combines this estimate with access recency, retaining recent contexts with higher continuation likelihood while gradually deprioritizing long-idle ones. 
We evaluate \textbf{Session-Aware} alone and with \textbf{Cache-Aware+Guard}. 
The appendix provides implementation details and policy parameters.

\paragraph{Workloads}
SCBench~\citep{li2025scbench} lacks request timestamps, so we construct a closed-loop multi-turn workload with Poisson session arrivals and sampled inter-turn think times, serving Qwen3-8B on one RTX~PRO~6000 with a 96-GiB L2 cache.
The production-derived Mooncake trace~\citep{qin2026mooncake} records request timestamps, input/output lengths, and remapped block hashes, preserving real arrivals and inter-turn gaps while exposing intra- and inter-session prefix sharing. We replay 50\%, 75\%, and 100\% of sessions beginning within a 15-minute window, preserving their start times and complete turn sequences. We serve Qwen3-8B on one H100 with a 96-GiB L2 cache.

\input{tables/main/tab_scbench}

\paragraph{SCBench Results}
Table~\ref{tab:scbench} shows trade-off among cache reuse, responsiveness, and batch completion. Compared with \textbf{FCFS}, which achieves only 19.2\% reuse due to prefix recomputation, \textbf{Cache-Aware} uses engine residency feedback to raise reuse to 87.6\%, reduce median TTFT from 63.1 to 4.6 seconds, and shorten batch completion from 481 to 224 seconds. Yet, repeatedly deferring cold requests increases maximum TTFT.
Waiting protection bounds this behavior. \textbf{Cache-Aware+Guard-40} improves P50, P95, and maximum TTFT over \textbf{FCFS} while retaining a 1.61$\times$ batch speedup. Increasing the threshold to 60 seconds favors reuse and batch completion but weakens tail-latency protection. 
These results exercise both directions of HEAR: engine feedback identifies reuse opportunities, while harness requirements constrain how aggressively the scheduler exploits them.

\begin{figure}[ht]
\centering
\includegraphics[width=0.99\linewidth]{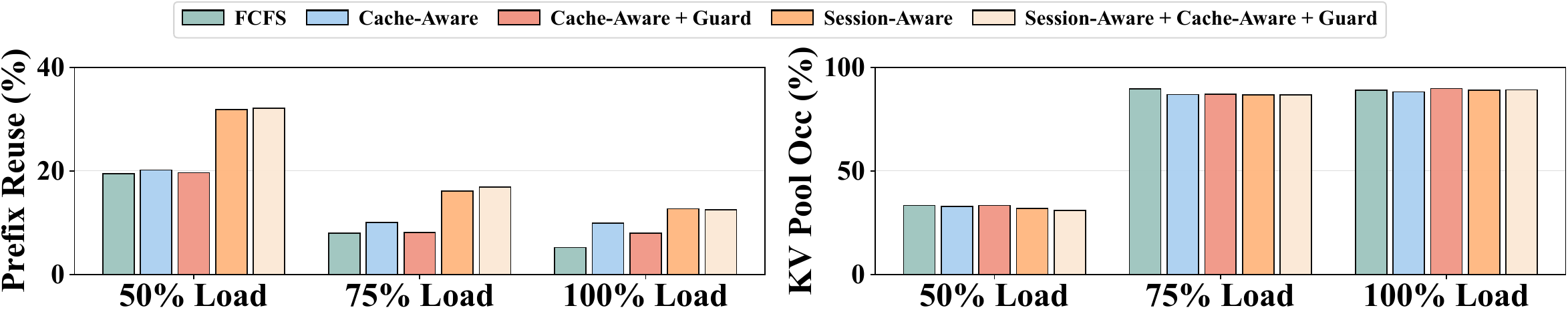}
\vspace{-0.5em}
\caption{\textbf{KV reuse and GPU KV occupancy on Mooncake Bench under different load levels.}}
\label{fig:cache}
\vspace{-0.5em}
\end{figure}

\paragraph{Mooncake Results}
Figure~\ref{fig:cache} and Table~\ref{tab:trace} show that the preferred policy varies with load while the HEAR interface remains unchanged. 
At 50\% load, limited queuing makes idle-period eviction the primary source of reuse loss; \textbf{Session-Aware} therefore outperforms execution-order control, raising reuse from 19.5\% to 31.9\%. 
At 75\%, high concurrency and remaining reuse opportunities make eviction- and execution-order control complementary: combining \textbf{Session-Aware} with \textbf{Cache-Aware + Guard} yields the shortest session time, highest throughput, and highest reuse. 
At 100\%, \textbf{Cache-Aware} achieves better P50, session latency, and throughput, whereas \textbf{Session-Aware} improves tail latency and reuse.
Appendix~\ref{app:run_timelines} provides execution timelines. 
Thus, no policy dominates across loads; HEAR enables workload-dependent use of workflow intent and engine feedback through unchanged message categories and interaction semantics.

\input{tables/main/tab_mooncake}

\subsection{Capability-Aware Execution Configuration}
\label{sec:capability_configuration}

\paragraph{HEAR Instantiation and Candidate Space}
We instantiate Equation~\ref{eq:coordination_policy} as an engine-side configuration selector. 
The harness description $h_i$ specifies the agent role, workflow stage, workload shape, execution objective, and quality requirements. 
The engine view $s_j$ provides compatible execution modes and their measured latency, throughput, KV-capacity, and stability profiles. 
These inputs correspond to HEAR's \emph{Execution Description and Intent}, \emph{Execution Requirements and Control}, and \emph{State and Capabilities}.
We consider the shared candidate space:
\begin{equation}
    \mathcal{C}
    =
    \{\mathrm{vLLM},\mathrm{OmniKV}\}
    \times
    \{\mathrm{vLLM},\mathrm{H_2O},\mathrm{SnapKV}\},
    \label{eq:capability_configuration_space}
\end{equation}
where the first and second sets denote the Main-side and Reader/sub-agent-side modes.  Among configurations passing development-stage quality and stability checks, the selector minimizes makespan:
\begin{equation}
    d_i^{\mathrm{config}}=\pi_{\mathrm{E}}(h_i,s_j),
    \qquad
    d_{\mathrm{BCP}}^{*}=(\mathrm{vLLM},\mathrm{H_2O}),
    \qquad
    d_{\mathrm{DRB}}^{*}=(\mathrm{OmniKV},\mathrm{H_2O}).
    \label{eq:selected_configurations}
\end{equation}
The mappings and method-specific parameters are verified and frozen
before formal evaluation, and are loaded by the Engine at workflow
initialization.

\paragraph{Workloads and Evaluation Design}
We evaluate all six configurations on 208 BrowseComp-Plus questions~\citep{chen2026browsecompplus} and 100 DeepResearchBench instances~\citep{du2026deepresearch} per arm, using identical within-benchmark budgets and independently initialized serving states. Both retain their native orchestration and retrieval environments within a common LangGraph-based Main--Retriever--Reader/sub-agent harness, with GLM-4.7-Flash serving Main and Reader/sub-agent inference on separate H100 GPUs.
Appendices~\ref{app:implementation} and~\ref{app:additional_results} detail selector calibration, execution-mode parameters, per-arm results, and quality and stability analyses.

\input{tables/main/brows}

\paragraph{End-to-End Results}
Table~\ref{tab:capability_configuration_results} reports end-to-end time, P95 latency, and task quality for all six configurations. Main entries are measured values, subscripts give ratios relative to vLLM--vLLM, and bold entries denote the frozen selections.
On BrowseComp-Plus, vLLM--H$_2$O reduces wall time from 5.69 to 4.62 hours, yielding a 1.23$\times$ speedup while preserving Judge accuracy (46.63\% vs. 46.15\%). 
On DeepResearchBench, OmniKV--H$_2$O reduces makespan from 5.93 to 2.42 hours, yielding a 2.45$\times$ speedup while preserving RACE (41.1\% vs. 40.6\%). 
Aggregate efficiency and P95 latency occasionally favor different mappings, confirming that selection must account for the execution objective communicated in $h_i$. 
Overall, both selected mappings improve end-to-end efficiency without an observed degradation in task quality.

\input{tables/main/agent_separate}

\paragraph{Role-Level Attribution}
Table~\ref{tab:role_level_attribution} compares configurations that change one role's mode while holding the other fixed. Reader/sub-agent execution provides the largest consistent gain. Selecting H$_2$O reduces mean latency by 5.60$\times$ on BrowseComp-Plus and 7.22$\times$ on DeepResearchBench. On BrowseComp-Plus, this gain accompanies a 76.2\% reduction in prefilled tokens and an increase in cache reuse from 6.2\% to 80.3\%; on DeepResearchBench, it accompanies 49.8\% chain reuse.
The preferred Main-side mode is workload-dependent. BrowseComp-Plus Main requests generate approximately 0.4K output tokens, and vLLM is $1.14\times$ faster than OmniKV. 
DeepResearchBench Main requests generate approximately 4.4K tokens in a round-synchronized workflow, and OmniKV is $1.08\times$ faster than vLLM. 
The longer decoding workload better amortizes mode-specific overhead. 

\paragraph{Fixed-Mapping Cost}
Neither selected mapping is optimal for both workloads. Applying OmniKV--H$_2$O universally would increase BrowseComp-Plus wall time by 7.8\%, whereas applying vLLM--H$_2$O universally would increase DeepResearchBench makespan by 14.0\%. 
Additional throughput--capacity and makespan--tail-latency trade-offs
are reported in Appendix~\ref{app:additional_results}.

\subsection{Cross-Policy Takeaway}
\label{sec:protocol_analysis}

The two policy families instantiate \textbf{HEAR} at different timescales: cache-aware coordination combines dynamic cache and queue state with harness intent and requirements to control request ordering and KV retention, while capability-aware configuration combines stable engine profiles with workload descriptions and objectives to select inference modes.
Both follow Equation~\ref{eq:coordination_policy}, since neither layer alone has sufficient information. Across conversational and research-agent workloads, decisions adapt to the workload while the HEAR interface and workflow semantics remain unchanged, demonstrating a common bidirectional contract for workload-specific policies.



%% file: tables/main/tab_scbench.tex
\begin{table}[!t]
\centering
\small
\setlength{\tabcolsep}{4pt}
\renewcommand{\arraystretch}{1.3}
\caption{\textbf{Cache-aware runtime coordination results on SCBench.} 
All times are in seconds. Batch completion is the elapsed time to complete all users' multi-turn conversations. Gray subscripts report speedups relative to \textbf{FCFS}. Reuse denotes the KV-cache reuse rate for follow-up prompt tokens.}
\label{tab:scbench}
\vspace{-0.5em}
\begin{tabular}{llllll}
\toprule
\multirow{2.5}{*}{\textbf{Configuration}} &
\multicolumn{3}{c}{\textbf{TTFT $\downarrow$}} &
\multirow{2.5}{*}[-0.4ex]{\textbf{Batch Completion $\downarrow$}} &
\multirow{2.5}{*}[-0.4ex]{\textbf{Reuse $\uparrow$}} \\
\cmidrule(lr){2-4}
& \textbf{P50} & \textbf{P95} & \textbf{MAX} & & \\
\midrule
\textbf{FCFS}
& 63.1$_{\color{gray}1.00\times}$
& 74.2$_{\color{gray}1.00\times}$
& 79.0$_{\color{gray}1.00\times}$
& 481$_{\color{gray}1.00\times}$
& 19.2 \\
\textbf{Cache-Aware}
& 4.6$_{\color{gray}13.6\times}$
& 75.8$_{\color{gray}0.98\times}$
& 93.9$_{\color{gray}0.84\times}$
& \textbf{224}$_{\color{black!70}\mathbf{2.15\times}}$
& \textbf{87.6} \\
\textbf{Cache-Aware + Guard-40}
& 28.3$_{\color{gray}2.23\times}$
& \textbf{51.1}$_{\color{black!70}\mathbf{1.45\times}}$
& \textbf{65.0}$_{\color{black!70}\mathbf{1.22\times}}$
& 299$_{\color{gray}1.61\times}$
& 66.1 \\
\textbf{Cache-Aware + Guard-60}
& \textbf{4.5}$_{\color{black!70}\mathbf{13.9\times}}$
& 64.0$_{\color{gray}1.16\times}$
& 80.4$_{\color{gray}0.98\times}$
& 230$_{\color{gray}2.09\times}$
& 84.7 \\
\bottomrule
\end{tabular}
\end{table}

%% file: tables/main/tab_mooncake.tex
\begin{table}[!t]
\centering
\small
\setlength{\tabcolsep}{3pt}
\renewcommand{\arraystretch}{1.3}
\caption{\textbf{Cache-aware runtime coordination results on Mooncake Bench.}
Session is the total response-waiting time per multi-turn conversation; throughput is measured in turns per minute.}
\label{tab:trace}
\vspace{-0.5em}
\resizebox{\columnwidth}{!}{%
\begin{tabular}{llllllll}
\toprule
\multirow{2}{*}[-0.4ex]{\textbf{Load}} &
\multirow{2}{*}[-0.4ex]{\textbf{Configuration}} &
\multicolumn{3}{c}{\textbf{TTFT $\downarrow$}} &
\multirow{2}{*}[-0.4ex]{\textbf{Session $\downarrow$}} &
\multirow{2}{*}[-0.4ex]{\textbf{Throughput $\uparrow$}} &
\multirow{2}{*}[-0.4ex]{\textbf{Reuse $\uparrow$}} \\
\cmidrule(lr){3-5}
& & \textbf{P50} & \textbf{P95} & \textbf{MAX} & & & \\
\midrule

\multirow{5}{*}{\textbf{50\%}}
& \textbf{FCFS}
& 0.7 {\color{gray}\scriptsize$\times$1.00}
& 2.8 {\color{gray}\scriptsize$\times$1.00}
& 6.2 {\color{gray}\scriptsize$\times$1.00}
& 13.7 {\color{gray}\scriptsize$\times$1.00}
& 86.1 {\color{gray}\scriptsize$\times$1.00}
& 19.5 \\

& \textbf{Cache-Aware}
& 0.6 {\color{gray}\scriptsize$\times$1.21}
& 2.4 {\color{gray}\scriptsize$\times$1.17}
& \textbf{5.9} {\color{black!70}\scriptsize$\mathbf{\times1.06}$}
& 13.3 {\color{gray}\scriptsize$\times$1.02}
& 86.0 {\color{gray}\scriptsize$\times$1.00}
& 20.2 \\

& \textbf{Cache-Aware + Guard}
& 0.6 {\color{gray}\scriptsize$\times$1.10}
& 2.8 {\color{gray}\scriptsize$\times$1.01}
& 7.4 {\color{gray}\scriptsize$\times$0.84}
& 13.6 {\color{gray}\scriptsize$\times$1.01}
& 86.2 {\color{gray}\scriptsize$\times$1.00}
& 19.6 \\

& \textbf{Session-Aware}
& 0.6 {\color{gray}\scriptsize$\times$1.16}
& 2.5 {\color{gray}\scriptsize$\times$1.14}
& 6.3 {\color{gray}\scriptsize$\times$0.99}
& 13.0 {\color{gray}\scriptsize$\times$1.05}
& 86.6 {\color{gray}\scriptsize$\times$1.01}
& 31.9 \\

& \textbf{Session-Aware + Cache-Aware + Guard}
& \textbf{0.5} {\color{black!70}\scriptsize$\mathbf{\times1.35}$}
& \textbf{2.3} {\color{black!70}\scriptsize$\mathbf{\times1.21}$}
& 6.1 {\color{gray}\scriptsize$\times$1.02}
& \textbf{12.7} {\color{black!70}\scriptsize$\mathbf{\times1.08}$}
& \textbf{86.8} {\color{black!70}\scriptsize$\mathbf{\times1.01}$}
& \textbf{32.1} \\

\midrule

\multirow{5}{*}{\textbf{75\%}}
& \textbf{FCFS}
& 10.9 {\color{gray}\scriptsize$\times$1.00}
& 51.9 {\color{gray}\scriptsize$\times$1.00}
& 56.7 {\color{gray}\scriptsize$\times$1.00}
& 50.2 {\color{gray}\scriptsize$\times$1.00}
& 116.4 {\color{gray}\scriptsize$\times$1.00}
& 8.0 \\

& \textbf{Cache-Aware}
& 8.7 {\color{gray}\scriptsize$\times$1.26}
& 42.9 {\color{gray}\scriptsize$\times$1.21}
& 49.6 {\color{gray}\scriptsize$\times$1.14}
& 42.6 {\color{gray}\scriptsize$\times$1.18}
& 119.2 {\color{gray}\scriptsize$\times$1.02}
& 10.1 \\

& \textbf{Cache-Aware + Guard}
& 9.1 {\color{gray}\scriptsize$\times$1.20}
& 42.3 {\color{gray}\scriptsize$\times$1.23}
& 47.9 {\color{gray}\scriptsize$\times$1.18}
& 42.5 {\color{gray}\scriptsize$\times$1.18}
& 119.3 {\color{gray}\scriptsize$\times$1.03}
& 8.1 \\

& \textbf{Session-Aware}
& 10.5 {\color{gray}\scriptsize$\times$1.04}
& \textbf{39.0} {\color{black!70}\scriptsize$\mathbf{\times1.33}$}
& \textbf{45.1} {\color{black!70}\scriptsize$\mathbf{\times1.26}$}
& 43.2 {\color{gray}\scriptsize$\times$1.16}
& 119.5 {\color{gray}\scriptsize$\times$1.03}
& 16.1 \\

& \textbf{Session-Aware + Cache-Aware + Guard}
& \textbf{8.2} {\color{black!70}\scriptsize$\mathbf{\times1.33}$}
& 39.8 {\color{gray}\scriptsize$\times$1.30}
& 45.4 {\color{gray}\scriptsize$\times$1.25}
& \textbf{39.8} {\color{black!70}\scriptsize$\mathbf{\times1.26}$}
& \textbf{121.0} {\color{black!70}\scriptsize$\mathbf{\times1.04}$}
& \textbf{16.9} \\

\midrule

\multirow{5}{*}{\textbf{100\%}}
& \textbf{FCFS}
& 163.9 {\color{gray}\scriptsize$\times$1.00}
& 277.0 {\color{gray}\scriptsize$\times$1.00}
& 281.5 {\color{gray}\scriptsize$\times$1.00}
& 256.9 {\color{gray}\scriptsize$\times$1.00}
& 122.7 {\color{gray}\scriptsize$\times$1.00}
& 5.2 \\

& \textbf{Cache-Aware}
& \textbf{86.8} {\color{black!70}\scriptsize$\mathbf{\times1.89}$}
& 209.9 {\color{gray}\scriptsize$\times$1.32}
& 220.9 {\color{gray}\scriptsize$\times$1.27}
& \textbf{156.1} {\color{black!70}\scriptsize$\mathbf{\times1.65}$}
& \textbf{128.3} {\color{black!70}\scriptsize$\mathbf{\times1.05}$}
& 10.0 \\

& \textbf{Cache-Aware + Guard}
& 97.1 {\color{gray}\scriptsize$\times$1.69}
& 209.2 {\color{gray}\scriptsize$\times$1.32}
& 217.6 {\color{gray}\scriptsize$\times$1.29}
& 178.7 {\color{gray}\scriptsize$\times$1.44}
& 122.8 {\color{gray}\scriptsize$\times$1.00}
& 8.0 \\

& \textbf{Session-Aware}
& 93.0 {\color{gray}\scriptsize$\times$1.76}
& \textbf{199.2} {\color{black!70}\scriptsize$\mathbf{\times1.39}$}
& \textbf{207.8} {\color{black!70}\scriptsize$\mathbf{\times1.35}$}
& 171.8 {\color{gray}\scriptsize$\times$1.50}
& 124.6 {\color{gray}\scriptsize$\times$1.02}
& \textbf{12.7} \\

& \textbf{Session-Aware + Cache-Aware + Guard}
& 161.7 {\color{gray}\scriptsize$\times$1.01}
& 267.4 {\color{gray}\scriptsize$\times$1.04}
& 277.7 {\color{gray}\scriptsize$\times$1.01}
& 243.2 {\color{gray}\scriptsize$\times$1.06}
& 108.9 {\color{gray}\scriptsize$\times$0.89}
& 12.5 \\

\bottomrule
\end{tabular}%
}
\end{table}

%% file: tables/main/brows.tex
\begin{table}[!t]
\centering
\small
\setlength{\tabcolsep}{4.0pt}
\renewcommand{\arraystretch}{1.3}
\caption{\textbf{Capability-aware execution configuration results.}}
\label{tab:capability_configuration_results}
\vspace{-0.5em}
\resizebox{0.99\textwidth}{!}{%
\begin{tabular}{lllll lll}
\toprule
\multirow{2.5}{*}{\textbf{Main}} &
\multirow{2.5}{*}{\textbf{Reader}} &
\multicolumn{3}{c}{\textbf{Browscomp-plus}} &
\multicolumn{3}{c}{\textbf{Deep-research-bench}} \\
\cmidrule(lr){3-5}
\cmidrule(lr){6-8}
&
&
\textbf{Wall (h) $\downarrow$} &
\textbf{P95 E2E (min) $\downarrow$} &
\textbf{Acc. (\%) $\uparrow$} &
\textbf{Makesp (h) $\downarrow$} &
\textbf{P95 E2E (min) $\downarrow$} &
\textbf{RACE (\%) $\uparrow$} \\
\midrule
\textbf{vLLM}
& \textbf{vLLM}
& $5.69_{\color{gray}1.00\times}$
& $40.0_{\color{gray}1.00\times}$
& 46.15
& $5.93_{\color{gray}1.00\times}$
& $39.8_{\color{gray}1.00\times}$
& 40.6 \\

\textbf{OmniKV}
& \textbf{vLLM}
& $5.37_{\color{gray}1.06\times}$
& $40.3_{\color{gray}0.99\times}$
& 47.12
& $6.17_{\color{gray}0.96\times}$
& $39.9_{\color{gray}1.00\times}$
& 41.1 \\

\textbf{vLLM}
& $\mathbf{H_2O}$
& $\mathbf{4.62}_{\color{black!70}\mathbf{1.23\times}}$
& $39.6_{\color{gray}1.01\times}$
& 46.63
& $2.76_{\color{gray}2.15\times}$
& $15.6_{\color{gray}2.55\times}$
& 40.5 \\

\textbf{OmniKV}
& $\mathbf{H_2O}$
& $4.98_{\color{gray}1.14\times}$
& $37.2_{\color{gray}1.08\times}$
& 43.27
& $\mathbf{2.42}_{\color{black!70}\mathbf{2.45\times}}$
& $15.8_{\color{gray}2.52\times}$
& 41.1 \\

\textbf{vLLM}
& \textbf{SnapKV}
& $4.83_{\color{gray}1.18\times}$
& $40.4_{\color{gray}0.99\times}$
& 44.23
& $2.85_{\color{gray}2.08\times}$
& $17.7_{\color{gray}2.25\times}$
& 40.0 \\

\textbf{OmniKV}
& \textbf{SnapKV}
& $5.29_{\color{gray}1.07\times}$
& $40.6_{\color{gray}0.99\times}$
& 43.75
& $2.76_{\color{gray}2.15\times}$
& $18.3_{\color{gray}2.17\times}$
& 40.5 \\

\bottomrule
\end{tabular}%
}
\end{table}

%% file: tables/main/agent_separate.tex
\begin{table}[!t]
\centering
\small
\renewcommand{\arraystretch}{1.3}
\caption{\textbf{Role-level execution attribution.}}
\label{tab:role_level_attribution}
\vspace{-0.5em}
\begin{tabular}{lllll}
\toprule
\multirow[l]{2}{*}[1.0em]{\textbf{Benchmark}}
& \shortstack{\textbf{Main Output}\\\textbf{Tok./Req.}}
& \shortstack{\textbf{Main Latency}\\
              \textbf{Alt. $\rightarrow$ Selected}}
& \shortstack{\textbf{Reader/Sub Latency}\\
              \textbf{Dense $\rightarrow$ Selected}}
& \multirow[l]{2}{*}[1.0em]{\textbf{Cache Evidence}} \\
\midrule
\textbf{BrowseComp-Plus}
& $\sim0.4$K
& \shortstack{
    OmniKV $\rightarrow$ vLLM\\
    $17.0\rightarrow14.9$ s
    $_{\color{gray}1.14\times}$
}
& \shortstack{
    vLLM $\rightarrow$ H$_2$O\\
    $53.4\rightarrow9.5$ s
    $_{\color{gray}5.60\times}$
}
& \shortstack{
    Prefill $-76.2\%$\\
    Reuse $6.2\%\rightarrow80.3\%$
} \\

\textbf{DeepResearchBench}
& $\sim4.4$K
& \shortstack{
    vLLM $\rightarrow$ OmniKV\\
    $69.4\rightarrow64.2$ s
    $_{\color{gray}1.08\times}$
}
& \shortstack{
    vLLM $\rightarrow$ H$_2$O\\
    $421.5\rightarrow58.4$ s
    $_{\color{gray}7.22\times}$
}
& Chain reuse $49.8\%$ \\

\bottomrule
\end{tabular}
\end{table}

%% file: 07_conclusions.tex
\section{Conclusions}
\label{sec:conclusion}

We introduce \textbf{HEAR}, a bidirectional Harness--Engine protocol that connects workflow intent and requirements with engine state, capabilities, and outcomes while preserving workflow and model semantics. 
We instantiate HEAR for cache-aware runtime coordination and role-specific inference configuration. 
On SCBench, waiting-protected cache-aware scheduling achieves a \textbf{1.61$\times$} batch speedup while improving median and tail TTFT. 
Mooncake further shows that future-use-aware retention must account for live queue state under overload. On BrowseComp-Plus and DeepResearchBench, workload-specific role configurations yield \textbf{1.23$\times$} and \textbf{2.45$\times$} end-to-end speedups, respectively, without observed degradation in task quality. 
These results establish HEAR as a reusable, policy-independent substrate for workload-adaptive coordination across the Harness-Engine boundary, with its benefits determined by the information and controls exposed by the serving stack.




\subsection*{AI use statement}

We used generative AI tools to assist with refining the conceptual
framework and experimental methodology, modifying and debugging code,
interpreting experimental results, revising parts of the manuscript,
Chinese--English translation, literature review, \LaTeX{} formatting,
and the preparation of scientific figures. The research questions,
system design, experimental decisions, and conclusions were determined
by the authors. Generative AI was not used to generate datasets or
experimental results. All AI-assisted content, technical claims,
figures, and citations were manually reviewed and verified by the
authors. The authors take full responsibility for the final content of
this work.

\subsection*{Ethics statement}

This work involves no human-subject experiments or collection of new
personal or sensitive data. Since the proposed protocol exchanges
execution and resource metadata, practical deployments should enforce
access control, tenant isolation, and bounded waiting. Our method
improves inference efficiency but does not address the safety or
reliability of the underlying model outputs.

\subsection*{Reproducibility statement}

The main text specifies the proposed protocol, policy instantiations,
workloads, baselines, evaluation metrics, and experimental settings.
Appendix~\ref{app:implementation} provides the implementation details
and execution-mode configurations needed to reproduce the reported
experiments.

%

%% file: 08_appendix.tex
\appendix


\section{Implementation Details}
\label{app:implementation}

\subsection{Protocol and Selector Implementation}
\label{app:selector_implementation}

\paragraph{Selector Inputs}
For capability-aware configuration, the Harness instantiates $h_i$
with the agent role, workflow stage, observed input and output workload
shape, concurrency and synchronization pattern, execution objective,
and applicable quality or feasibility requirements.
The Engine instantiates $s_j$ with the execution modes supported for
that role and their development profiles, including request latency,
throughput, KV-cache reuse and capacity behavior, recoverable capacity
retries, and failure or fallback status.
Each profile is associated with a concrete Engine configuration and
source revision.

\paragraph{Development Calibration and Verification}
A calibration script first filters configurations that are
incompatible with the role or model, have incomplete or invalid runs,
or fail the development quality and stability checks.
Let $\mathcal{C}_{\mathrm{valid}}$ denote the remaining candidates.
The script ranks them by measured development makespan:
\begin{equation}
    \widetilde{d}
    =
    \arg\min_{d\in\mathcal{C}_{\mathrm{valid}}}
    T_{\mathrm{dev}}(d),
    \label{eq:development_selector}
\end{equation}
where $\widetilde{d}$ is the programmatically recommended mapping.
The recommendation is then verified against the execution logs and
the quality and stability summaries.
The verifier may select another candidate from
$\mathcal{C}_{\mathrm{valid}}$ only when these checks expose an invalid
measurement, an unsupported execution path, or a feasibility or
reliability issue not captured by makespan.
The accepted mapping, denoted by $d^*$, is serialized before formal
evaluation.
Formal-evaluation results are not used during ranking or verification.

\paragraph{Freezing and Runtime Application.}
The accepted role-to-mode mapping, method-specific parameters, and
Engine revisions are serialized in a versioned configuration file.
At workflow initialization, the Engine-side selector loads this file
and resolves each role to the corresponding preconfigured Engine
service; it does not search the six-arm space for every request.
This procedure freezes
\begin{equation}
    d_{\mathrm{BCP}}^{*}
    =
    (\mathrm{vLLM},\mathrm{H_2O}),
    \qquad
    d_{\mathrm{DRB}}^{*}
    =
    (\mathrm{OmniKV},\mathrm{H_2O}).
    \label{eq:app_frozen_configurations}
\end{equation}
The runtime records the requested and actual execution modes together
with completion, failure, or fallback outcomes.
Neither the capability profiles, ranking rule, nor frozen mappings are
updated using the formal six-arm results.

\subsection{Cache-Aware Runtime Coordination Settings}
\label{app:cache_coordination}

\paragraph{Benchmark Construction and Fair Replay}
For both benchmarks, the formal evaluation follows a trajectory recording
and replay procedure.
We first execute and record a complete trajectory, including request
arrival times, input and output lengths, session/turn relationships, and
prefix information.
All configurations then replay the same trajectory.
This ensures that different coordination policies do not observe
different workloads due to changes in generated content, input/output
lengths, or subsequent execution trajectories; configurations differ
only in request scheduling and KV-cache management behavior.

For SCBench~\citep{li2025scbench}, the original data contain multi-turn
interactions but do not provide real user request timestamps.
We construct each sample as an independent session and use a Poisson
process to simulate initial session arrivals.
After each response completes, a user think time is sampled to determine
when the next turn of the same session enters the system.
Think time follows
\[
\mathcal{N}(5,1)\ \mathrm{s}.
\]
After the trajectory is recorded once, all configurations replay the
same session-arrival and think-time sequences.

For Mooncake Bench~\citep{qin2026mooncake}, we use the request
timestamps, input and output lengths, and block-hash information recorded
in the production trace, while preserving the original session start
times, inter-turn intervals, and complete turn sequences.
Different load levels are constructed only by replaying 50\%, 75\%, and
100\% of the sessions whose first requests fall within the same
15-minute window, without altering the original timing relationships
between requests.

\paragraph{Experimental Parameters}
SCBench uses Qwen3-8B \cite{qwen3technicalreport} on a single RTX~PRO~6000 with a 96-GiB L2 KV
cache.
Mooncake Bench also uses Qwen3-8B, running on a single H100 with a
96-GiB L2 KV cache.
For SCBench, the Guard waiting-protection threshold is evaluated at
$W=40$\,s and $W=60$\,s; for Mooncake Bench, we use $W=40$\,s.
Except for the coordination policy being compared, all experiments use
the same replay trajectory, model configuration, cache capacity, and
other serving parameters.

\paragraph{Coordination and Session-Aware KV Retention}
\textsc{Cache-Aware} orders ready requests according to the KV-residency
view maintained by the harness.
For each request, the harness matches reusable prefix blocks
sequentially from the beginning of the prompt.
Matching stops at the first missing or evicted block, since subsequent
blocks cannot independently form a reusable prefix.
The harness separately counts reusable prefix tokens in the GPU and CPU
tiers.
A request is classified as \emph{hot} if reusable prefixes cover at
least 50\% of the prompt, and \emph{cold} otherwise.

Cold requests retain the default priority, while the priority of a hot
request is defined as the negative number of reusable KV blocks:
\[
\mathrm{priority}
=
-N_{\mathrm{reuse}},
\]
where $N_{\mathrm{reuse}}$ denotes the number of reusable KV blocks.
Because the engine schedules requests with smaller priority values first,
hot requests are executed before cold requests, and hot requests are
further ordered by decreasing reusable-prefix length.

The residency view reflects the cache state currently observed by the
harness and does not constitute a residency guarantee; when a request
actually executes, the engine rechecks the reusable content against the
current cache state.

\textsc{Guard} is submitted by the harness to the engine at fixed times
and supports subsequent updates.
Once a request has waited longer than the threshold $W$, the engine
promotes it to a protected priority so that later cache-hot requests can
no longer bypass it.
Protected requests are still ordered according to their waiting order.

\textsc{Session-Aware} is independent of the request-scheduling priority
above and acts only on KV-cache eviction order.
The harness maintains a session-return probability $p$ by continuously
summarizing completed conversations and their subsequent return
behavior, and uses this statistic to provide a continuation estimate for
the current session.
The estimate is updated only from completed conversations and does not
use future information from the current replay.

After a request completes, the engine records the corresponding
context's $p$ and completion time and computes its retention priority as
\[
r
=
p \cdot e^{-\mathrm{idle}/400}.
\]
Here, $\mathrm{idle}$ denotes the time elapsed since the context's most
recent request completed.
Contexts that are more likely to return and have been accessed more
recently receive higher retention priority, while the priority gradually
decays as idle time increases.

\subsection{Protocol Cost Accounting}
\label{app:protocol_overhead}

HEAR adds control-plane metadata processing rather than model-side
computation.
It does not introduce additional model layers or per-token GPU
operators; its online work consists of state updates, object lookups,
policy-input construction, control handling, and execution-outcome
reporting.
The computation introduced by OmniKV, H$_2$O, or SnapKV belongs to the
corresponding execution mode rather than to the protocol itself.

For cache-aware coordination, Engine KV events update a
Harness-maintained residency directory.
At each scheduling point, the Harness queries reusable prefix blocks,
constructs scheduling preferences, and attaches the applicable waiting
requirement.
The Engine applies these controls and reports the realized execution
order, cache reuse, retries, and errors.
For capability-aware configuration, candidate profiling, filtering,
ranking, and verification occur during offline development.
The online path loads the frozen role-to-mode mapping once at workflow
initialization and performs a constant-size lookup when dispatching
each model request.

Table~\ref{tab:protocol_cost_accounting} summarizes the frequency and
cost attribution of these operations.
All formal HEAR-enabled runs include the corresponding online
control-plane path; the end-to-end results in the main text therefore
reflect the net performance after these operations.

\input{tables/appendix/cost}

\subsection{Role-Specific Models and Execution Modes}
\label{app:execution_mode_configurations}

All configurations use GLM-4.7-Flash \cite{5team2025glm45agenticreasoningcoding}, with Main-side and
Reader/sub-agent-side inference placed on separate H100 GPUs.
Dense execution uses vLLM~0.26.0, while OmniKV, H$_2$O, and SnapKV
are provided through SparseEngine.
Tables~\ref{tab:bcp_execution_mode_configurations}
and~\ref{tab:drb_execution_mode_configurations} summarize the
cache, capacity, and mode-specific settings used in the formal
BrowseComp-Plus and DeepResearchBench runs, respectively.
We report the behavior-affecting parameters here; complete launch
commands, environment manifests, and source hashes are included in
the supplementary artifact.

\input{tables/appendix/brows_config}

For BrowseComp-Plus, the Harness allows at most four Main and twelve
Reader requests in flight; generation uses temperature zero,
top-$p=1$, and disables model thinking.
OmniKV is based on SparseEngine revision
\texttt{67174b1b}; the exact patched source snapshots are archived with
the artifact.

\input{tables/appendix/deepres_config}

For DeepResearchBench, Main requests use temperature $0.2$,
top-$p=0.9$, a 4,096-token output limit, and model thinking.
Sub-agent requests use the same sampling parameters with a
512-token output limit and disable thinking.
All sparse modes use SparseEngine revision \texttt{469aa791}.

\paragraph{SnapKV Prefill Admission}
SnapKV uses a compressed per-layer retention budget of 4,672 tokens,
but does not sparsify the prefill phase.
Consequently, a long request temporarily reserves space for its full
prompt until prefill completes.
For DeepResearchBench, the Harness therefore applies a FIFO prefill
gate to SnapKV sub-agent requests, limiting the global in-flight
prefill workload to 456,837 tokens.
A grant is released upon the first streamed output token, when the
Engine clears the request's full-prefill reservation.
This Harness-side control is separate from SnapKV's Engine-side
retention parameters.

\section{Additional Results}
\label{app:additional_results}

\subsection{Runtime Behavior of Cache-Aware Runtime Coordination}
\label{app:run_timelines}

\begin{figure*}[t]
    \centering
    \includegraphics[width=0.98\textwidth]{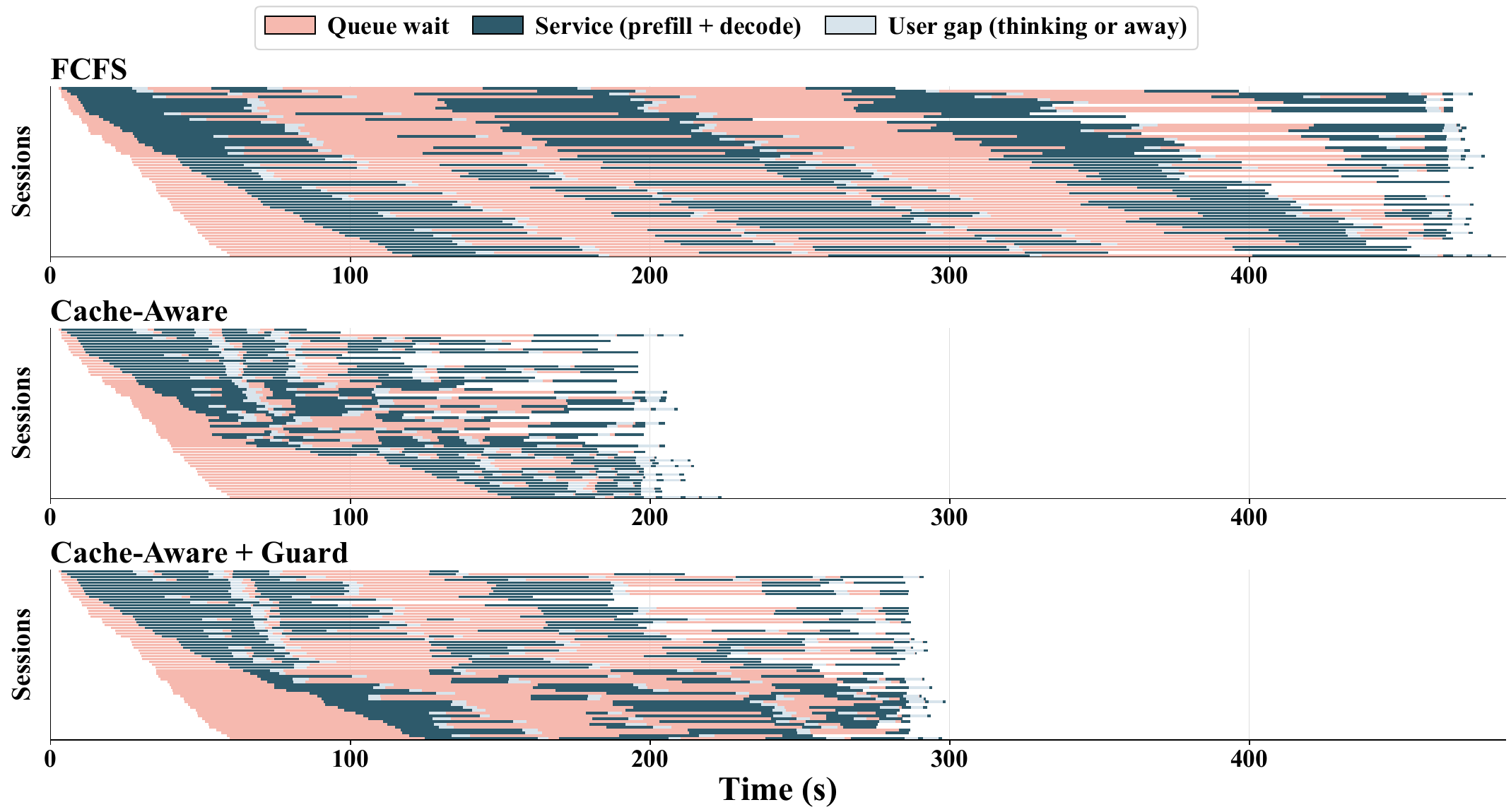}
    \caption{\textbf{Runtime execution timelines on SCBench.}
    Each session is decomposed into queue wait, model service
    (prefill + decode), and user gap.}
    \label{fig:scbench_timeline}
\end{figure*}

Figure~\ref{fig:scbench_timeline} shows the runtime timelines of the
closed-loop SCBench workload.
Each turn is decomposed into queue wait, model service, and user gap.
Because all configurations replay the same session-arrival and
think-time trajectory, the user-gap pattern is shared across methods,
and the main differences arise from request execution order.
Under \textsc{FCFS}, queueing accumulates and substantially extends the
overall execution horizon.
\textsc{Cache-Aware} prioritizes requests with reusable prefixes and
concentrates more service earlier in the trace.
Adding \textsc{Guard} limits prolonged waiting caused by repeated
bypass while retaining the cache-aware execution-order policy.

\begin{figure*}[!t]
\centering
\includegraphics[width=0.99\textwidth]{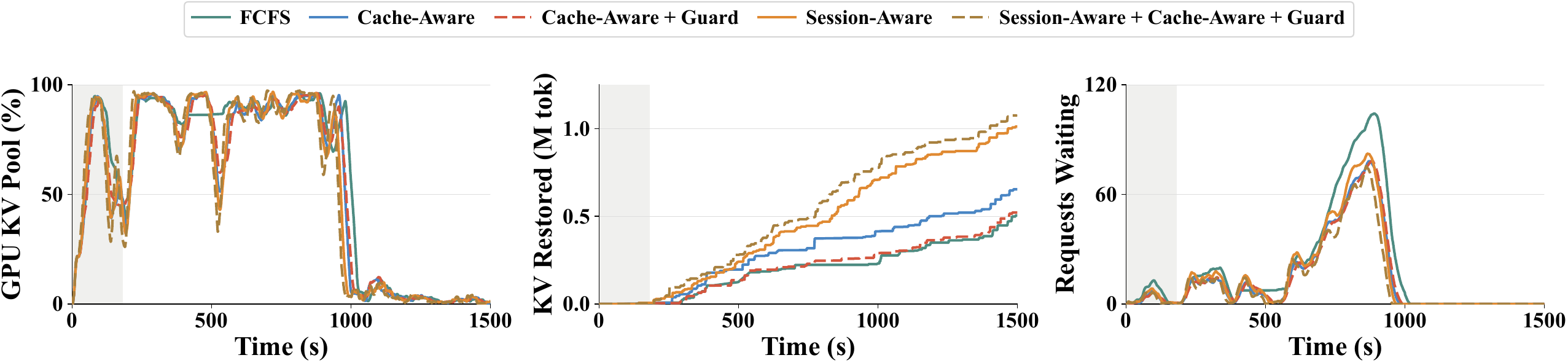}
\vspace{-0.5em}
\caption{\textbf{Runtime behavior under the Mooncake replay.}
GPU KV-pool occupancy, cumulative KV restoration, and the number of waiting requests over time for different scheduling policies.}
\label{fig:mooncake_load}
\vspace{-0.5em}
\end{figure*}

Figure~\ref{fig:mooncake_load} shows the runtime behavior of the different
scheduling policies under the Mooncake replay. GPU KV occupancy follows a
similar trajectory across policies, while the main differences appear in KV
restoration and queuing: session-aware variants restore substantially more KV,
whereas cache-aware scheduling keeps restoration lower. Compared with FCFS, the
coordinated policies also reduce the peak number of waiting requests, with Guard
further limiting queue buildup without materially changing the overall
KV-occupancy pattern.

\subsection{Complete Six-Arm Results of Role-specific inference configuration}
\label{app:full_six_arm_results}

Table~\ref{tab:appendix_browsecomp_full} reports the complete
BrowseComp-Plus results over the shared $2\times3$ execution-mode
space.
Panel~(a) includes aggregate efficiency and per-question latency,
whereas Panel~(b) reports task quality and completion outcomes.
Throughput is measured in questions per hour, and all end-to-end
latencies are reported in minutes.
C, I, U, and F denote correct, incorrect, unresolved, and run-failure
questions, respectively.

Table~\ref{tab:appendix_drb_full} reports the corresponding
DeepResearchBench results.
Panel~(a) includes makespan, throughput, and per-instance latency,
while Panel~(b) reports the overall RACE score, its four quality
dimensions, and completion outcomes.
All quality scores are multiplied by 100.
$N$ denotes the number of reports evaluated by the RACE judge;
Comp., Instr., and Read. denote comprehensiveness, instruction
following, and readability, respectively.

\input{tables/appendix/brow_full_result}
\input{tables/appendix/deepres_full_result}

As shown in Tables~\ref{tab:appendix_browsecomp_full}
and~\ref{tab:appendix_drb_full}, the configurations selected during
development achieve the lowest aggregate completion time within their
respective benchmarks.
BrowseComp-Plus selects vLLM--H$_2$O, while DeepResearchBench selects
OmniKV--H$_2$O.
In both cases, the efficiency improvement is obtained without an
observed degradation in task quality.
The complete results also expose secondary trade-offs: for example,
OmniKV--H$_2$O yields a lower BrowseComp-Plus P95 latency despite not
minimizing its wall time, while the DeepResearchBench sparse
configurations differ substantially more in execution efficiency than
in RACE.

\subsection{Role-Level Workload and Execution Analysis}
\label{app:role_level_results}

The two benchmarks share the same high-level role organization but
expose substantially different workload shapes.
Table~\ref{tab:appendix_role_workload} reports the role-level workload
observed under the configuration selected for each benchmark.
Values are normalized per benchmark item, i.e., per question for
BrowseComp-Plus and per instance for DeepResearchBench.

\input{tables/appendix/workload_shape}
\input{tables/appendix/execution_behavior}

\subsection{Quality Evaluation}
\label{app:quality_evaluation}

\paragraph{BrowseComp-Plus}
We evaluate answer quality over all 208 questions in each arm.
Unresolved questions and run failures remain in the accuracy
denominator, and all formal Judge labels are available.
Table~\ref{tab:app_bcp_answer_quality} reports the complete outcome
distribution and Wilson 95\% confidence intervals, while
Table~\ref{tab:app_bcp_evidence_quality} reports retrieval and citation
quality.

\input{tables/appendix/brows_answer_quality}
\input{tables/appendix/brows_retrieve_quality}

\paragraph{DeepResearchBench}
We evaluate generated reports using the official reference-free RACE
evaluator.
Because task errors and empty reports reduce the number of valid
articles available to the Judge, Table~\ref{tab:app_drb_quality}
reports both evaluation coverage and the four RACE dimensions.

\input{tables/appendix/deepres_answer_quality}

\paragraph{Paired Quality Comparisons}
Table~\ref{tab:app_paired_quality} compares each benchmark's selected
configuration with its all-vLLM baseline.
BrowseComp-Plus uses all 208 paired questions.
For DeepResearchBench, we use the 89 instances that receive valid RACE
scores in every arm, ensuring a common paired cohort for the six-arm
analysis.
Confidence intervals are percentile bootstrap intervals over paired
questions or instances.

\input{tables/appendix/paired_quality_comparisons}

\paragraph{Judge Provenance}
All BrowseComp-Plus outputs are evaluated with Qwen3-32B using a
4,096-token Judge output limit and the same intended evaluation
settings.
Arms A, B, and F are judged on an RTX~PRO~6000, whereas Arms C, D, and
E are judged on an H100.
We retain this provenance and report per-question paired outcomes;
re-evaluating all arms in a single Judge environment would further
remove this environmental difference.
DeepResearchBench uses the vendored official RACE evaluator with
gpt-5.4-mini.
Provider errors from the initial high-concurrency Judge pass are
re-evaluated at a concurrency of four.
The reported tables contain the final successful Judge outputs.

%% file: tables/appendix/cost.tex
\begin{table}[!t]
\centering
\small
\setlength{\tabcolsep}{4pt}
\renewcommand{\arraystretch}{1.3}
\caption{\textbf{HEAR protocol cost accounting.}}
\label{tab:protocol_cost_accounting}
\vspace{-0.5em}
\begin{tabular}{
    >{\raggedright\arraybackslash}m{0.17\textwidth}
    >{\raggedright\arraybackslash}m{0.20\textwidth}
    >{\raggedright\arraybackslash}m{\dimexpr0.45\textwidth-6\tabcolsep\relax}
    >{\raggedright\arraybackslash}m{0.15\textwidth}
}
\toprule
\textbf{Component}
& \textbf{Frequency}
& \textbf{Additional Online Work}
& \textbf{Included in Formal E2E Time} \\
\midrule

KV-state update
& Per cache event
& Residency-directory metadata update
& Yes \\

Scheduling decision
& Per dispatch point
& Prefix lookup, priority construction, and Guard check
& Yes \\

Role-to-mode lookup
& Per model request
& Lookup in the frozen role mapping
& Yes \\

Outcome reporting
& Per inference/control outcome
& Status, actual mode, reuse, retry, and error metadata
& Yes \\

Profile calibration
& Offline development stage
& Candidate profiling, filtering, ranking, and verification
& No \\

\bottomrule
\end{tabular}
\end{table}

%% file: tables/appendix/brows_config.tex
\begin{table}[!t]
\centering
\small
\setlength{\tabcolsep}{4pt}
\renewcommand{\arraystretch}{1.3}
\caption{\textbf{Execution-mode configurations for BrowseComp-Plus.}}
\label{tab:bcp_execution_mode_configurations}
\vspace{-0.5em}
\begin{tabular}{
    >{\raggedright\arraybackslash}m{0.095\textwidth}
    >{\raggedright\arraybackslash}m{0.245\textwidth}
    >{\raggedright\arraybackslash}m{0.190\textwidth}
    >{\raggedright\arraybackslash}m{0.390\textwidth}
}
\toprule

\textbf{Mode}
& \textbf{Cache and Token Settings}
& \textbf{Execution Capacity}
& \textbf{Mode-specific Parameters}
\\

\midrule

\rowcolor{black!6}
\multicolumn{4}{c}{\textbf{Main Role}}
\\

\textbf{vLLM}
& \textbf{Prefix caching}\newline
  Max.\ context 202,752;\newline
  32,768 batched tokens
& 16 sequences
& \textbf{Dense execution}
\\

\addlinespace[2pt]

\textbf{OmniKV}
& \textbf{Radix cache}\newline
  32,768-token prefill chunks
& 4 batch / 4 decode /\newline
  16 GPU sequences
& \textbf{Full-attention layers:} automatic;\newline
  \textbf{Retention:}
  8 sink / 256 recent / 4,096 decode tokens;\newline
  \textbf{Prefill workspace:} 8 GiB
\\

\midrule

\rowcolor{black!6}
\multicolumn{4}{c}{\textbf{Reader Role}}
\\

\textbf{vLLM}
& \textbf{Prefix caching}\newline
  Max.\ context 202,752;\newline
  16,384 batched tokens
& 16 sequences
& \textbf{Dense execution}
\\

\addlinespace[2pt]

\textbf{H$_2$O}
& \textbf{Chain cache}\newline
  16,384-token prefill chunks
& 12 batch / 12 decode /\newline
  36 GPU sequences
& \textbf{Budgets:}
  16,384 prefill / 8,192 decode tokens;\newline
  \textbf{Recent ratio:} 0.2;\newline
  \textbf{Scoring:}
  128-token window with logits-based prefill scoring
\\

\addlinespace[2pt]

\textbf{SnapKV}
& \textbf{Chain cache}\newline
  16,384-token prefill chunks
& 12 batch / 12 decode /\newline
  36 GPU sequences
& \textbf{Retention:}
  64 sink / 512 recent / 4,096 decode tokens;\newline
  \textbf{Observation window:} 32 tokens;\newline
  \textbf{Other:}
  no full-attention layers or prefill sparsification
\\

\bottomrule
\end{tabular}

\end{table}

%% file: tables/appendix/deepres_config.tex
\begin{table}[!t]
\centering
\small
\setlength{\tabcolsep}{4pt}
\renewcommand{\arraystretch}{1.3}
\caption{\textbf{Execution-mode configurations for DeepResearchBench.}}
\label{tab:drb_execution_mode_configurations}
\vspace{-0.5em}
\begin{tabular}{
    >{\raggedright\arraybackslash}m{0.095\textwidth}
    >{\raggedright\arraybackslash}m{0.245\textwidth}
    >{\raggedright\arraybackslash}m{0.195\textwidth}
    >{\raggedright\arraybackslash}m{0.385\textwidth}
}
\toprule

\textbf{Mode}
& \textbf{Cache and Token Settings}
& \textbf{Execution Capacity}
& \textbf{Mode-specific Parameters}
\\

\midrule

\rowcolor{black!6}
\multicolumn{4}{c}{\textbf{Main Role}}
\\

\textbf{vLLM}
& \textbf{Prefix caching}\newline
  Max.\ context 200,000; 32,768 batched tokens
& 10 sequences
& \textbf{Dense execution}
\\

\addlinespace[2pt]

\textbf{OmniKV}
& \textbf{Radix cache}\newline
  32,768-token prefill chunks
& 10 batch / 10 decode / 16 GPU sequences
& \textbf{Full-attention layers:}
  $\{0, 3, 9, 18, 22, 27, 43\}$
\\

\midrule

\rowcolor{black!6}
\multicolumn{4}{c}{\textbf{Sub-Agent Role}}
\\

\textbf{vLLM}
& \textbf{Prefix caching}\newline
  Max.\ context 200,000; 32,768 batched tokens
& 10 sequences
& \textbf{Dense execution}
\\

\addlinespace[2pt]

\textbf{H$_2$O}
& \textbf{Chain cache}\newline
  32,768-token prefill chunks
& 10 batch / 10 decode / 44 GPU sequences
& \textbf{Budgets:}
  8,192 prefill / 4,096 decode tokens;\newline
  \textbf{Recent ratio:} 0.2;\newline
  \textbf{Scoring:}
  128-token window with logits-based prefill scoring
\\

\addlinespace[2pt]

\textbf{SnapKV}
& \textbf{Chain cache}\newline
  32,768-token prefill chunks
& 10 batch / 10 decode / 44 GPU sequences
& \textbf{Retention:}
  64 sink / 512 recent / 4,096 decode tokens
  (4,672 total);\newline
  \textbf{Observation window:} 32 tokens;\newline
  \textbf{Other:}
  no full-attention layers or prefill sparsification
\\

\bottomrule
\end{tabular}

\end{table}

%% file: tables/appendix/brow_full_result.tex
\begin{table}[!t]
\centering
\small
\setlength{\tabcolsep}{2pt}
\renewcommand{\arraystretch}{1.5}
\caption{\textbf{Complete BrowseComp-Plus six-arm results.}}
\label{tab:appendix_browsecomp_full}
\vspace{-0.5em}
\resizebox{\textwidth}{!}{%
\begin{tabular}{ll cccccc cccccc}
\toprule
\multirow[c]{2.5}{*}{\textbf{Main}}
& \multirow[c]{2.5}{*}{\textbf{Reader}} 
& \multicolumn{6}{c}{\textbf{\textit{Efficiency and Latency}}}
& \multicolumn{6}{c}{\textbf{\textit{Quality and Completion Outcomes}}} \\
\cmidrule(lr){3-8}
\cmidrule(lr){9-14}
& 
& \multicolumn{1}{l}{\textbf{Wall (h) $\downarrow$}}
& \multicolumn{1}{l}{\textbf{Speedup $\uparrow$}}
& \multicolumn{1}{l}{\textbf{Throughput $\uparrow$}}
& \multicolumn{1}{l}{\textbf{Median $\downarrow$}}
& \multicolumn{1}{l}{\textbf{P95 $\downarrow$}}
& \multicolumn{1}{l}{\textbf{Max $\downarrow$}}
& \multicolumn{1}{l}{\textbf{Acc. (\%) $\uparrow$}}
& \multicolumn{1}{l}{\textbf{Retr. Recall (\%) $\uparrow$}}
& \multicolumn{1}{l}{\textbf{C}}
& \multicolumn{1}{l}{\textbf{I}}
& \multicolumn{1}{l}{\textbf{U}}
& \multicolumn{1}{l}{\textbf{F}} \\
\midrule

\multirow[c]{3}{*}{\textbf{vLLM}}
& \textbf{vLLM}
& 5.69
& $1.00\times$
& 36.57
& 14.63
& 40.00
& 41.32
& 46.15
& 69.22
& 96
& 46
& 54
& 12 \\

& \textbf{H$_2$O}
& \textbf{4.62}
& $\mathbf{1.23\times}$
& \textbf{45.00}
& \textbf{10.98}
& \textbf{39.58}
& \textbf{41.81}
& \textbf{46.63}
& \textbf{70.17}
& \textbf{97}
& \textbf{59}
& \textbf{45}
& \textbf{7} \\

& \textbf{SnapKV}
& 4.83
& $1.18\times$
& 43.03
& 10.75
& 40.38
& 43.26
& 44.23
& 70.56
& 92
& 63
& 40
& 13 \\

\midrule

\multirow[c]{3}{*}{\textbf{OmniKV}}
& \textbf{vLLM}
& 5.37
& $1.06\times$
& 38.72
& 13.67
& 40.28
& 46.54
& 47.12
& 68.75
& 98
& 42
& 53
& 15 \\

& \textbf{H$_2$O}
& 4.98
& $1.14\times$
& 41.73
& 12.64
& 37.24
& 42.37
& 43.27
& 68.33
& 90
& 61
& 49
& 8 \\

& \textbf{SnapKV}
& 5.29
& $1.07\times$
& 39.28
& 11.14
& 40.60
& 42.60
& 43.75
& 69.46
& 91
& 56
& 43
& 18 \\

\bottomrule
\end{tabular}%
}

\end{table}

%% file: tables/appendix/deepres_full_result.tex
\begin{table}[!t]
\centering
\small
\setlength{\tabcolsep}{2pt}
\renewcommand{\arraystretch}{1.5}
\caption{\textbf{Complete DeepResearchBench six-arm results.}}
\label{tab:appendix_drb_full}
\vspace{-0.5em}
\resizebox{\textwidth}{!}{%
\begin{tabular}{ll ccc cccc cccc cccc}
\toprule
\multirow[l]{4}{*}{\textbf{Main}}
& \multirow[l]{4}{*}{\textbf{Sub-Agent}}
& \multicolumn{7}{c}{\textbf{\textit{Efficiency and Latency}}}
& \multicolumn{8}{c}{\textbf{\textit{Quality and Completion Outcomes}}} \\
\cmidrule(lr){3-9}
\cmidrule(lr){10-17}
&
& \multirow[c]{2}{*}[-0.5ex]{
    \textbf{Makespan $\downarrow$}
}
& \multirow[c]{2}{*}[-0.5ex]{
    \textbf{Speedup $\uparrow$}
}
& \multirow[c]{2}{*}[-0.5ex]{
    \textbf{Throughput $\uparrow$}
}
& \multicolumn{4}{c}{
    \textbf{E2E Latency (min) $\downarrow$}
}
& \multirow[c]{2}{*}{\textbf{$N$}}
& \multirow[c]{2}{*}[-0.5ex]{
     \textbf{RACE (\%) $\uparrow$}
}
& \multirow[c]{2}{*}{\textbf{Comp.}}
& \multirow[c]{2}{*}{\textbf{Insight}}
& \multirow[c]{2}{*}{\textbf{Instr.}}
& \multirow[c]{2}{*}{\textbf{Read.}}
& \multirow[c]{2}{*}{\textbf{Errors}}
& \multirow[c]{2}{*}{\textbf{Empty}} \\

\noalign{\vskip -1pt}
\cmidrule(lr){6-9}
\noalign{\vskip -1pt}

&
&
&
&
&
\textbf{Mean}
& \textbf{Median}
& \textbf{P95}
& \textbf{Max}
&
&
&
&
&
&
&
& \\

\midrule

\multirow[l]{3}{*}{\textbf{vLLM}}
& \textbf{vLLM}
& 5.93
& $1.00\times$
& 16.7
& 29.46
& 30.64
& 39.81
& 41.18
& 97
& 40.65
& 39.93
& 38.68
& 42.98
& 42.91
& 1
& 4 \\

& $\mathbf{H_2O}$
& 2.76
& $2.15\times$
& 36.3
& 13.17
& 13.17
& 15.64
& 21.27
& 96
& 40.46
& 39.65
& 38.68
& 42.81
& 42.57
& 0
& 4 \\

& \textbf{SnapKV}
& 2.85
& $2.08\times$
& 35.1
& 15.13
& 15.19
& 17.70
& 20.26
& 99
& 40.01
& 39.33
& 37.78
& 42.51
& 42.45
& 0
& 1 \\

\midrule

\multirow[l]{3}{*}{\textbf{OmniKV}}
& \textbf{vLLM}
& 6.17
& $0.96\times$
& 16.2
& 30.41
& 30.64
& 39.93
& 40.64
& 98
& 41.14
& 40.30
& 39.26
& 43.57
& 43.20
& 0
& 2 \\

& $\mathbf{H_2O}$
& \textbf{2.42}
& $\mathbf{2.45\times}$
& \textbf{41.3}
& \textbf{12.21}
& \textbf{12.24}
& \textbf{15.82}
& \textbf{17.06}
& \textbf{99}
& \textbf{41.05}
& \textbf{39.91}
& \textbf{39.59}
& \textbf{43.26}
& \textbf{43.10}
& \textbf{0}
& \textbf{1} \\

& \textbf{SnapKV}
& 2.76
& $2.15\times$
& 36.3
& 14.70
& 14.66
& 18.33
& 19.63
& 99
& 40.50
& 39.44
& 38.62
& 42.99
& 42.83
& 0
& 1 \\

\bottomrule
\end{tabular}%
}
\end{table}

%% file: tables/appendix/workload_shape.tex
\begin{table}[!t]
\centering
\small
\setlength{\tabcolsep}{4pt}
\renewcommand{\arraystretch}{1.5}
\caption{\textbf{Role-level workload shapes under the selected configurations.}}
\label{tab:appendix_role_workload}
\vspace{-0.5em}
\resizebox{\textwidth}{!}{%
\begin{tabular}{llllll}
\toprule
\textbf{Benchmark}
& \textbf{Role}
& \textbf{Req./Item}
& \textbf{Input/Req. (K)}
& \textbf{Output/Req. (K)}
& \textbf{Workflow Pattern}
\\
\midrule
\multirow[l]{2}{*}{\textbf{BrowseComp-Plus}}
& \textbf{Main}
& 3.0
& 52.5
& 0.36
& Dynamic coordination and answer synthesis
\\
& \textbf{Reader}
& 13.3
& 59.1
& 0.36
& Question-dependent document reads
\\
\midrule
\multirow[l]{2}{*}{\textbf{DeepResearchBench}}
& \textbf{Main}
& 5.4
& 32.5
& 4.42
& Round-synchronized review and synthesis
\\
& \textbf{Sub-Agent}
& 12.0
& 135.3
& 0.51
& Four parallel agents over three fixed rounds
\\
\bottomrule
\end{tabular}}
\end{table}

%% file: tables/appendix/execution_behavior.tex
\begin{table}[!t]
\centering
\small
\setlength{\tabcolsep}{4pt}
\renewcommand{\arraystretch}{1.3}
\caption{\textbf{Role-level execution behavior under arm-pair comparisons.}}
\label{tab:appendix_role_execution}
\vspace{-0.5em}
\begin{tabular}{llllll}
\toprule
\textbf{Benchmark}
& \textbf{Role}
& \textbf{Alternative $\rightarrow$ Selected}
& \textbf{Mean Latency (s)}
& \textbf{Gain}
& \textbf{Execution Evidence} \\
\midrule

\multirow[c]{2}{*}[-1.2ex]{\textbf{BrowseComp-Plus}}
& \textbf{Main}
& OmniKV $\rightarrow$ vLLM
& $17.0 \rightarrow 14.9$
& $1.14\times$
& \makecell[l]{
    Selected Main output:\\
    0.36K tokens/request
} \\

& \textbf{Reader}
& vLLM $\rightarrow$ H$_2$O
& $53.4 \rightarrow 9.5$
& $5.60\times$
& \makecell[l]{
    Prefill $-76.2\%$;\\
    reuse $6.2\% \rightarrow 80.3\%$
} \\

\midrule

\multirow[c]{2}{*}[-1.2ex]{\textbf{DeepResearchBench}}
& \textbf{Main}
& vLLM $\rightarrow$ OmniKV
& $69.4 \rightarrow 64.2$
& $1.08\times$
& \makecell[l]{
    Selected Main output:\\
    4.42K tokens/request
} \\

& \textbf{Sub-Agent}
& vLLM $\rightarrow$ H$_2$O
& $421.5 \rightarrow 58.4$
& $7.22\times$
& \makecell[l]{
    Chain reuse:\\
    $49.8\%$
} \\

\bottomrule
\end{tabular}
\end{table}

%% file: tables/appendix/brows_answer_quality.tex
\begin{table}[!t]
\centering
\small
\setlength{\tabcolsep}{4pt}
\renewcommand{\arraystretch}{1.3}
\caption{\textbf{BrowseComp-Plus answer-quality outcomes.}}
\label{tab:app_bcp_answer_quality}
\vspace{-0.5em}
\begin{tabular}{ll cccccc}
\toprule
\multirow[c]{2.5}{*}{\textbf{Main}}
& \multirow[c]{2.5}{*}{\textbf{Reader}} 
& \multicolumn{4}{c}{\textbf{Outcome Counts}}
& \multicolumn{2}{c}{\textbf{Answer Quality}} \\
\cmidrule(lr){3-6}
\cmidrule(lr){7-8}
&
& \textbf{Correct}
& \textbf{Incorrect}
& \textbf{Unresolved}
& \shortstack[l]{\textbf{Failure}}
& \shortstack[l]{\textbf{Acc. (\%)}}
& \shortstack[l]{\textbf{95\% CI}} \\
\midrule

\textbf{vLLM}
& \textbf{vLLM}
& 96 & 46 & 54 & 12
& 46.15
& $[39.51,\,52.94]$ \\

\textbf{OmniKV}
& \textbf{vLLM}
& 98 & 42 & 53 & 15
& 47.12
& $[40.45,\,53.89]$ \\

\textbf{vLLM}
& $\mathbf{H_2O}$
& 97 & 59 & 45 & 7
& 46.63
& $[39.98,\,53.41]$ \\

\textbf{OmniKV}
& $\mathbf{H_2O}$
& 90 & 61 & 49 & 8
& 43.27
& $[36.72,\,50.06]$ \\

\textbf{vLLM}
& \textbf{SnapKV}
& 92 & 63 & 40 & 13
& 44.23
& $[37.65,\,51.02]$ \\

\textbf{OmniKV}
& \textbf{SnapKV}
& 91 & 56 & 43 & 18
& 43.75
& $[37.18,\,50.54]$ \\

\bottomrule
\end{tabular}
\end{table}

%% file: tables/appendix/brows_retrieve_quality.tex
\begin{table}[!t]
\centering
\small
\setlength{\tabcolsep}{5pt}
\renewcommand{\arraystretch}{1.3}
\caption{\textbf{BrowseComp-Plus retrieval and citation quality.}}
\label{tab:app_bcp_evidence_quality}
\vspace{-0.5em}
\begin{tabular}{ll cccc}
\toprule
\multirow[c]{2.5}{*}{\textbf{Main}}
& \multirow[c]{2.5}{*}{\textbf{Reader}}
& \multirow[c]{2.5}{*}{\textbf{Answers Produced}}
& \multicolumn{3}{c}{\textbf{Evidence Quality (\%)}} \\
\cmidrule(lr){4-6}
&
&
& \textbf{Retrieval Recall $\uparrow$}
& \textbf{Citation Precision $\uparrow$}
& \textbf{Citation Recall $\uparrow$} \\
\midrule

\textbf{vLLM}
& \textbf{vLLM}
& 142 & 69.22 & 78.01 & 30.82 \\

\textbf{OmniKV}
& \textbf{vLLM}
& 140 & 68.75 & 77.71 & 33.90 \\

\textbf{vLLM}
& $\mathbf{H_2O}$
& 156 & 70.17 & 75.44 & 28.62 \\

\textbf{OmniKV}
& $\mathbf{H_2O}$
& 151 & 68.33 & 74.78 & 27.87 \\

\textbf{vLLM}
& \textbf{SnapKV}
& 155 & 70.56 & 72.63 & 28.10 \\

\textbf{OmniKV}
& \textbf{SnapKV}
& 147 & 69.46 & 75.03 & 28.47 \\

\bottomrule
\end{tabular}
\end{table}

%% file: tables/appendix/deepres_answer_quality.tex
\begin{table}[!t]
\centering
\small
\setlength{\tabcolsep}{4pt}
\renewcommand{\arraystretch}{1.3}
\caption{\textbf{DeepResearchBench RACE quality and evaluation coverage.}}
\label{tab:app_drb_quality}
\vspace{-0.5em}
\begin{tabular}{ll ccc ccccc}
\toprule
\multirow[c]{2.5}{*}{\textbf{Main}}
& \multirow[c]{2.5}{*}{\textbf{Sub-Agent}}
& \multicolumn{3}{c}{\textbf{Evaluation Coverage}}
& \multicolumn{5}{c}{\textbf{RACE (\%)}} \\
\cmidrule(lr){3-5}
\cmidrule(lr){6-10}
&
& \textbf{Judged}
& \textbf{Empty}
& \textbf{Task Errors}
& \textbf{Overall}
& \textbf{Comp.}
& \textbf{Insight}
& \textbf{Instr Follow.}
& \textbf{Read.} \\
\midrule

\textbf{vLLM}
& \textbf{vLLM}
& 97 & 4 & 1
& 40.65 & 39.93 & 38.68 & 42.98 & 42.91 \\

\textbf{OmniKV}
& \textbf{vLLM}
& 98 & 2 & 0
& 41.14 & 40.30 & 39.26 & 43.57 & 43.20 \\

\textbf{vLLM}
& $\mathbf{H_2O}$
& 96 & 4 & 0
& 40.46 & 39.65 & 38.68 & 42.81 & 42.57 \\

\textbf{OmniKV}
& $\mathbf{H_2O}$
& 99 & 1 & 0
& 41.05 & 39.91 & 39.59 & 43.26 & 43.10 \\

\textbf{vLLM}
& \textbf{SnapKV}
& 99 & 1 & 0
& 40.01 & 39.33 & 37.78 & 42.51 & 42.45 \\

\textbf{OmniKV}
& \textbf{SnapKV}
& 99 & 1 & 0
& 40.50 & 39.44 & 38.62 & 42.99 & 42.83 \\

\bottomrule
\end{tabular}
\end{table}

%% file: tables/appendix/paired_quality_comparisons.tex
\begin{table}[!t]
\centering
\small
\setlength{\tabcolsep}{4pt}
\renewcommand{\arraystretch}{1.4}
\caption{\textbf{Paired quality comparison against all-vLLM.}}
\label{tab:app_paired_quality}
\vspace{-0.5em}
\begin{tabular}{llllll}
\toprule
\textbf{Benchmark}
& \textbf{Selected}
& \textbf{Paired $N$}
& \textbf{Quality Diff. (\%)}
& \textbf{95\% Bootstrap CI (\%)}
& \textbf{Paired Test}
\\
\midrule

\textbf{BrowseComp-Plus}
& vLLM--H$_2$O
& 208
& $+0.48$
& $[-5.29,\,6.25]$
& \shortstack[l]{
    McNemar\\
    $p=1.000$
}
\\


\textbf{DeepResearchBench}
& OmniKV--H$_2$O
& 89
& $+0.51$
& $[-0.36,\,1.46]$
& \shortstack[l]{
    Paired $t$-test\\
    $p=.278$
}
\\

\bottomrule
\end{tabular}

\end{table}